\documentclass[journal]{IEEEtran}

\IEEEoverridecommandlockouts                              

\usepackage[ruled]{algorithm2e} 

\SetAlFnt{\small}
\SetAlCapFnt{\small}
\SetAlCapNameFnt{\small}
\SetAlCapHSkip{0pt}
\IncMargin{-\parindent} 

\usepackage{float}
\usepackage{multicol}
\usepackage{multirow}
\usepackage{algpseudocode}
\usepackage{color}
\usepackage{xcolor}
\usepackage{graphicx}
\usepackage{bm}
\usepackage{amsmath}
\usepackage{amssymb}
\usepackage{tabularx}
\usepackage{subcaption}
\usepackage{url}
\usepackage{cite}
\usepackage{prettyref}
\usepackage{booktabs}
\usepackage{siunitx}
\usepackage{makecell}
\usepackage{comment}
\usepackage{wrapfig}
\usepackage{hyperref}

\graphicspath{
    {./fig/}
}
    
\title{Spring-IMU Fusion Based Proprioception for Feedback Control of Soft Manipulators}

\author{Yinan Meng, Guoxin Fang$^{\dagger}$,~\IEEEmembership{Member,~IEEE}, Jiong Yang, Yuhu Guo, 
\\ and Charlie C.L. Wang$^{\dagger}$,~\IEEEmembership{Senior Member,~IEEE}

\thanks{The project is partially supported by the chair professorship fund of the University of Manchester and the research fund of UK Engineering and Physical Sciences Research Council (EPSRC) (Ref.\#: EP/W024985/1).}
\thanks{All authors are with the School of Engineering, The University
of Manchester, M13 9PL Manchester, U.K. }
\thanks{$^{\dagger}$Corresponding authors: \textit{Guoxin Fang and Charlie C.L. Wang}}
}

\usepackage{color}
\usepackage[normalem]{ulem}

\newcommand{\charlie}[1]{\textbf{\textcolor[rgb]{0.80,0.00,0.00}{[Charlie::~#1]}}}
\newcommand{\guoxin}[1]{\textcolor[rgb]{0.20,0.20,1.00}{[Guoxin::~#1]}}
\newcommand{\Yinan}[1]{\textcolor[rgb]{0.50,0.80,0.00}{[Yinan::~#1]}}

\newcommand{\revise}[2]{{}{#2}}

\definecolor{Rosolic}{cmyk}{0.00,1.00,0.50,0}
\definecolor{bleudefrance}{rgb}{0.19, 0.55, 0.91}
\definecolor{green}{cmyk}{0.3,0.2,0.95,0.0}
\definecolor{brown}{cmyk}{0.4,0.7,1.0,0.5}
\definecolor{dblue}{cmyk}{1,0.97,0.35,0.0}
\definecolor{Cyan}{cmyk}{1,0,0,0}
\definecolor{Azure}{rgb}{0.0, 0.5, 1.0}

\begin{document}

\maketitle

\begin{abstract}
This paper presents a novel framework to realize proprioception and closed-loop control for soft manipulators. Deformations with large elongation and large bending can be precisely predicted using geometry-based sensor signals obtained from the inductive springs and the \textit{inertial measurement units} (IMUs) with the help of machine learning techniques. Multiple geometric signals are fused into robust pose estimations, and a data-efficient training process is achieved after applying the strategy of sim-to-real transfer. As a result, we can achieve proprioception that is robust to the variation of external loading and has an average error of 0.7\% across the workspace on a pneumatic-driven soft manipulator.
The realized proprioception on soft manipulator is then contributed to building a sensor-space based algorithm for closed-loop control. A gradient descent solver is developed to drive the end-effector to achieve the required poses by iteratively computing a sequence of reference sensor signals. A conventional controller is employed in the inner loop of our algorithm to update actuators (i.e., the pressures in chambers) for approaching a reference signal in the sensor-space. The systematic function of closed-loop control has been demonstrated in tasks like path following and pick-and-place under different external loads. 
%
\end{abstract}



\section{Introduction}\label{secIntro}

\IEEEPARstart{S}{oft} manipulators present the promising ability to perform complex deformation with its compliance to safely interact with environments~\cite{falkenhahn2016dynamic, hughes2016soft}. They are now playing important roles in the automation of both industrial and medical applications (e.g.,~\cite{ranzani2015bioinspired, gong2019opposite}). Compared with their rigid counterparts, high \textit{degrees-of-freedom} (DOFs) in the configuration of soft bodies brings higher complexity in realizing effective modeling, shape sensing, and control~\cite{Nature2015}. When external loading or interaction is applied to the system, kinematic/dynamic models based on analytical formulations~\cite{della2020model, huang2021kinematic} or numerical simulations~\cite{fang2020kinematics, goury2018fast} are hard to precisely predict the deformation for accomplishing tasks like pick-and-place. 

\begin{figure}[t]
\centering
\includegraphics[width=\linewidth]{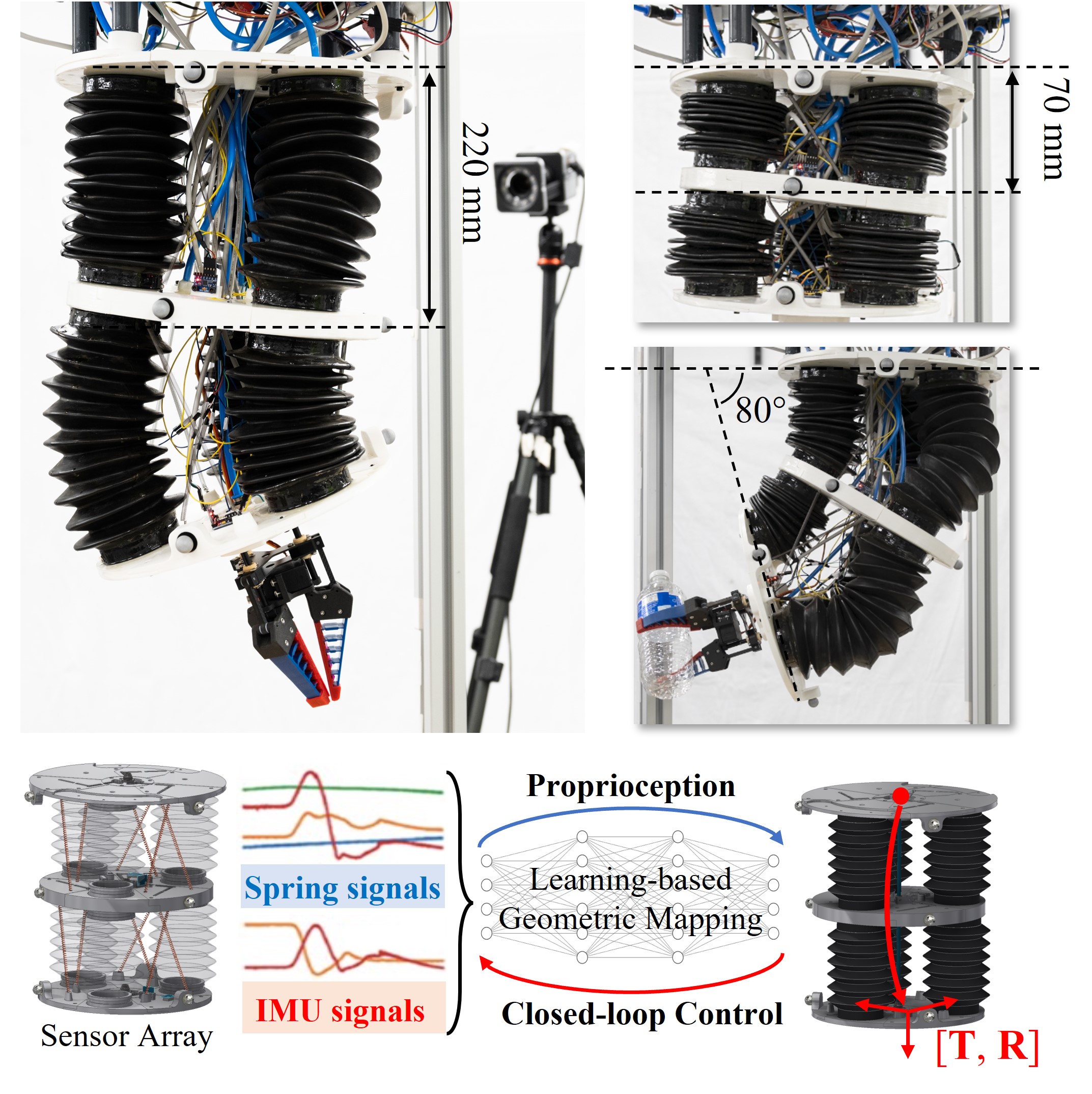}
\caption{Robust proprioception with high precision can be achieved by our method on a soft manipulator that performs deformations with large elongation and bending. The sensor signals collected from conductive springs and IMUs are fused and processed by a learning-based method to effectively predict the poses of the end-effector (i.e., the translation $\mathbf{T}$ and the rotation $\mathbf{R}$). A closed-loop controller through sensor space can then be developed to effectively complete tasks in different scenarios.
}\label{fig:teaser}
\end{figure}

A variety of sensing methods~\cite{case2016sensor, zhao2021shape, hughes2021sensing} have been employed to realize the proprioception of soft robots in tasks with closed-loop control. However, \revise{loading-independent proprioception and control for deformations with }{effective sensor systems for soft manipulators under large elongation and bending is still an unsolved problem (ref.~\cite{falkenhahn2016dynamic,wang2018toward}).} 
\revise{A new method is proposed in this paper to tackle this challenge of proprioception. When the sensor signals are only related to the deformation but not the external loading, this enables an effective closed-loop control for scenarios with different external loads. It is a challenging task to find a proper sensor to support the shape prediction of soft manipulators with large deformation. An example is given in Fig.~\ref{fig:teaser}, where the chambers of a soft robotic arm can extend to $3 \times$ of its initial length and bend up to $80^{\circ}$.}{Specifically,}
large elongation of chambers prevents the usage of bending sensors (such as optical fibers employed in~\cite{zhao2016optoelectronically, galloway2019fiber} or the strain gauges~\cite{elgeneidy2018directly, navarro2020model}) that can only work in a small range of elasticity. 
Although conductive spring was introduced by Sahu \textit{et al.}~\cite{sahu2022spring} as a solution to detect distance change on a soft robot, springs have less ability to capture the change of orientation on the end-effector when pure bending deformation is present~\cite{alcaide2017design,sahu2022spring}.
On the other aspect, large bending and buckling behavior can generate vision obstacles may lead to failure proprioception~\cite{hofer2021vision} when using vision pattern or lighting/color-based sensing techniques~\cite{3D-Deformation-sensing, hofer2021vision, scharff2022rapid}. 
In these cases where soft manipulators perform large changes in pose, sensors that are able to capture the global postures are more effective - such as IMUs~\cite{hughes2021sensing} or magnetic-field based sensors~\cite{baaij2023learning}. 
In order to capture the full posture of soft robots instead of only orientations, Josie \textit{et al.}~\cite{hughes2021sensing} presented a method to compensate sensor signals with a kinematics model based on \textit{piecewise constant curvature} (PCC)~\cite{della2020improved}. However, combining sensor signals with kinematics models can hardly maintain accuracy when external loading is applied~\cite{zhao2021shape,felt2019inductance}.
\revise{}{On the other hand,}
Navarro \textit{et al.}~\cite{navarro2020model} combined vision and tactile sensor data to achieve shape and force prediction of soft robots. Similarly, the direct combination of images and signals from soft strain sensors has been reported in~\cite{thuruthel2022multimodel}. 
\revise{non-vision based method (with or without sensor fusion) for sensing deformation in large elongation and bending does not exist in literature. }{
However, vision signals can be lost easily
in 
environments with numerous visual obstacles~\cite{fang2019vision}.}

In this paper, we introduce a sensing system that contains conductive springs and IMUs to achieve effective and stable proprioception of soft manipulators in large deformation. The sensor signals from conductive springs and IMUs are \revise{first processed to}{} geometrically-defined features, including the lengths of the springs (calibrated from inductance) and the rotation angles of the rigid base holding bellows (directly obtained from IMU). We hypothesize and prove from experiments that using these geometry-based sensor fusion can take advantage of different types of signals, and therefore outperform the existing solutions (the comparison is illustrated in Sec.~\ref{subsec:sensorSelection}). \revise{}{Meanwhile, geometry-oriented signals can maintain their independence from external loads, which enables stable proprioception for control.}
\revise{, a learning pipeline based on \textit{feedforward neural network} (FNN) is used}{We adopted learning-based method~\cite{thuruthel2018model,zhao2021shape,scharff2022rapid, brena2020choosing, qiu2022multi}} to analyze and encode \revise{those }{}geometrically defined long-range features (detail presented in Sec.~\ref{secLearning}). The learning outcome provides a high-precision mapping between the sensor signals and the poses of the end effector.  
The sim-to-real transfer strategy is employed to avoid directly learning the end-to-end mappings entirely from physical data, 
\revise{}{which may cause long data generation time and the (possible) drifted mechanical properties of elastomers on soft manipulators~\cite{fang2022efficient,baaij2023learning}.} Specifically, the network is first trained from a large dataset generated from a geometric-based simulator~\cite{fang2020kinematics}. 
\revise{Therefore, this}{ After that, the} reality gap is eliminated by training a lightweight sim-to-real network with a small physical dataset (ref.~\cite{bern2020soft,fang2022efficient}).

On the other hand, controlling a soft manipulator to reach target poses meets challenges including the nature of under-actuation \cite{wang2022control}, the hyper-elasticity of bellows \cite{Nature2015}, and the non-linearity in actuation system~\cite{george2018control}. Building an accurate closed-loop controller (e.g., model-based \cite{hughes2021sensing, goury2018fast} or learning-based~\cite{thuruthel2018model}) relies on the precision of proprioception to provide soft robot poses as feedback signals. \revise{For the control tasks where external loads are applied, loading-independent proprioception, as the main property of the sensor system introduced in this work, is needed. }{Taking advantage of the loading-independent proprioception realized by spring-IMU sensor fusion}, we introduce a sensor-space-based controller to achieve closed-loop control using sensor feedback. \revise{}{This two-level controller includes} a gradient-descent solver akin to~\cite{fang2020kinematics} \revise{is first applied and run in the outer loop of the algorithm. In each iteration of this outer loop, }{run in the outer loop to compute} a target sensor status as a reference based on the current status of the manipulator in the sensor space. A conventional controller~\cite{han2009pid} is adopted in the inner loop to adjust the pressures of chambers to reach the outer loop specified sensor signals as reference. Iteratively updating the references in the outer loop will drive the soft manipulator to reach a target pose\revise{. The efficiency of the proposed control algorithm is supported by the highly efficient computation for evaluating the sensor-pose mapping function and the gradient of control objective by the network's forward/backward propagation.}{, which can be implemented efficiently by the network's forward/backward propagation.}
\revise{The proposed control paradigm through sensor space is detailed}{Details can be found} in Sec.~\ref{secControl}.

The technical contributions of our work are as follows:
\begin{itemize}
\item A loading-independent sensing system containing springs and IMUs is introduced to realize precise proprioception of soft manipulators under deformations with large elongation and large bending.

\item A data-efficient training process is developed for establishing the sensor-pose mapping by only requiring a small dataset acquired on the physical setup with the help of sim-to-real learning transfer.

\item A sensor-space based algorithm with a high converging speed is developed to realize the function of closed-loop control under different external loads.
\end{itemize}
The behavior of the proposed sensor-based proprioception is verified on a laboratory-made soft manipulator. Loading-independent and accurate proprioception -- with an average predicted translation error of $0.7\%$ across the manipulator's workspace is achieved even when large external loads up to 500 grams are applied. The effectiveness of using this proprioception technique in closed-loop control has been demonstrated in tasks like path following and pick-and-place. \revise{Quantitative analysis on the convergence and precision is presented in Sec.~\ref{secResult}. }{}To the best of our knowledge, this is the first time a systematic approach is presented to address the issues of loading-independent motion control for soft manipulators.

\section{Soft Manipulator with Sensors}\label{secSetup}
The proposed proprioception and closed-loop control are conducted on a laboratory-made soft manipulator equipped with conductive springs and IMUs as sensors. Hardware of the soft manipulator is first presented in this section, which is followed by introducing the sensing system. Lastly, the method for calibrating sensor signals is detailed. 

\begin{figure}[t]
\includegraphics[width=\linewidth]{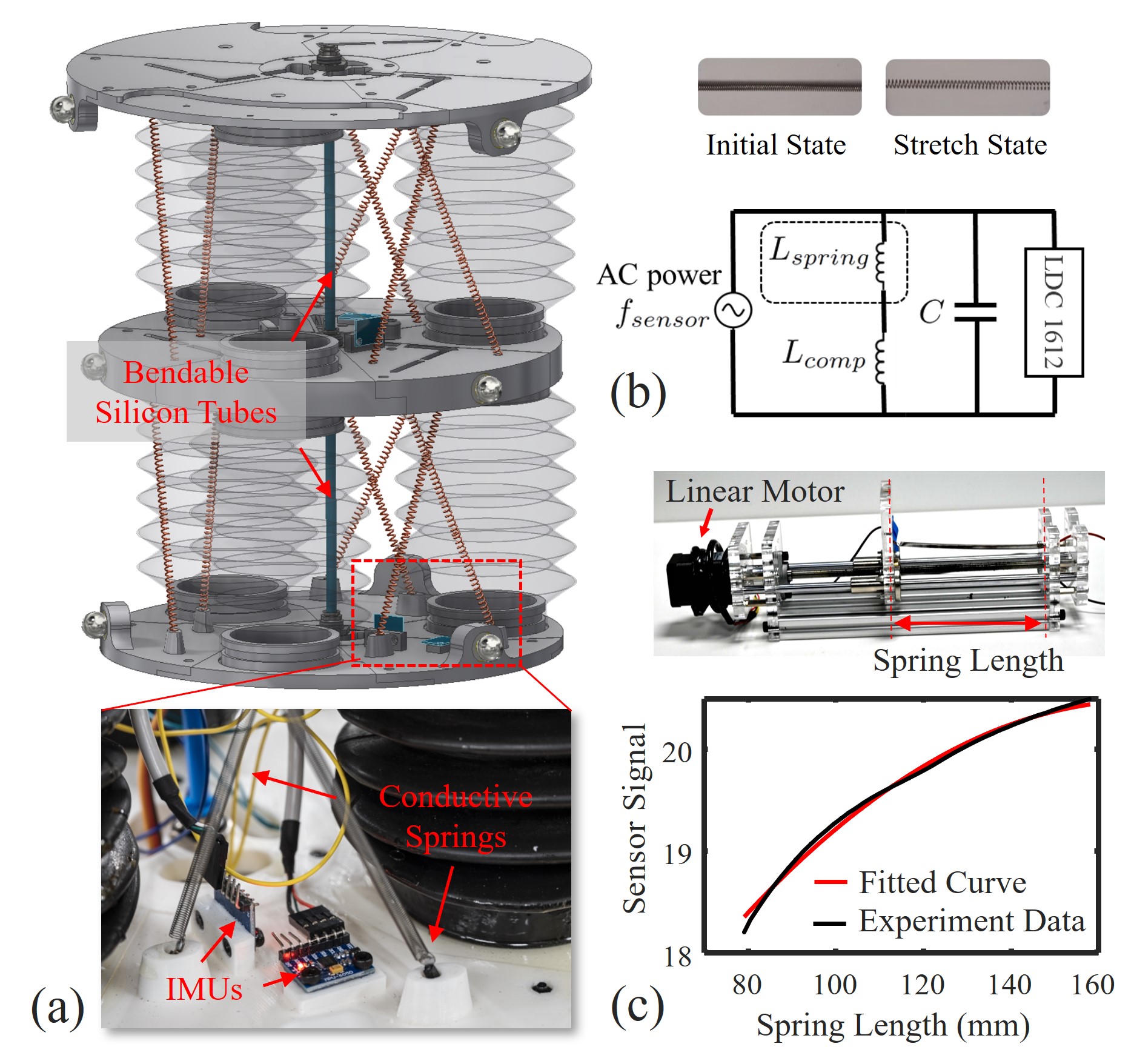}
\caption{(a) The illustration of sensors installed on a soft manipulator with two segments. Each segment contains six conductive spring sensors (shown in brown) and two IMUs (shown in blue). (b) The electronic circuit used to read the inductance of springs is given in the top-right corner. \revise{}{(c) Calibration of spring sensor to build the mapping from signal to spring length by a quartic polynomial function.}
}\label{fig:setup}
\end{figure}

\subsection{Hardware of Soft Manipulator}\label{subsec:hardware}
A \revise{}{pneumatic-driven} soft manipulator is fabricated in our work using six off-the-shelf rubber bellows as illustrated in Fig.~\ref{fig:teaser}. 
\revise{This is an extension of the previous design published in~\cite{3D-Deformation-sensing}. Different from \cite{3D-Deformation-sensing}, a ceiling-mounted design is adopted. }{We chose the bellow model V6-00400 from Freudenberg with an accordion structure. Each bellow with the diameters 65mm (inner) / 100mm (outer) and the 1mm thickness can be extended and retracted in the range of [70, 220]~mm under the pressure of [-40, 40]~kPa.} 
The soft manipulator consists of two segments, where each contains three bellows connected in parallel \revise{}{to enable a workspace of $300\times300\times300~\mathrm{mm}$. The entire soft manipulator system can achieve a bending angle of up to 80 degrees. }
\revise{Deformations with large elongation and large bending can be performed on the manipulator when bellows are pneumatically actuated. We use micro pumps (i.e., DEWIN DC 12V Micro Vacuum Pump) to control the flow rates of both}{For each chamber, a pair of micropumps is used to manage the} the inlet and outlet airflow via the pulse-width modulation of the voltage. \revise{}{We selected the micropump with an airflow rate of 1.8 L/min and implemented a closed-loop controller with pressure feedback (details in Sec.~\ref{secControl}). This ensures stable pressure to rapidly drive the system and compensates for minor leakages.}
To avoid the twisting behavior happening due to a relatively low stiffness of the bellow, a bendable silicon tube is used to connect the rigid pads located at the two ends of each segment (see the illustration in Fig.~\ref{fig:setup}). 
The bottom rigid pad of the lower segment is considered as the base of the end-effector, the status of which is described by a translation vector $\mathbf{T} = [x, y, z]$ and a rotation matrix $\mathbf{R}$ in our system. The soft manipulator is fitted with a soft gripper weighing 0.45 kg at the base, serving as the end-effector for grasping. \revise{}{Total weight of the soft manipulator is $1.97~\mathrm{kg}$} and objects up to $0.6~\mathrm{kg}$ can be picked up in our experimental tests.  





\begin{figure}[t]
\includegraphics[width=\linewidth]{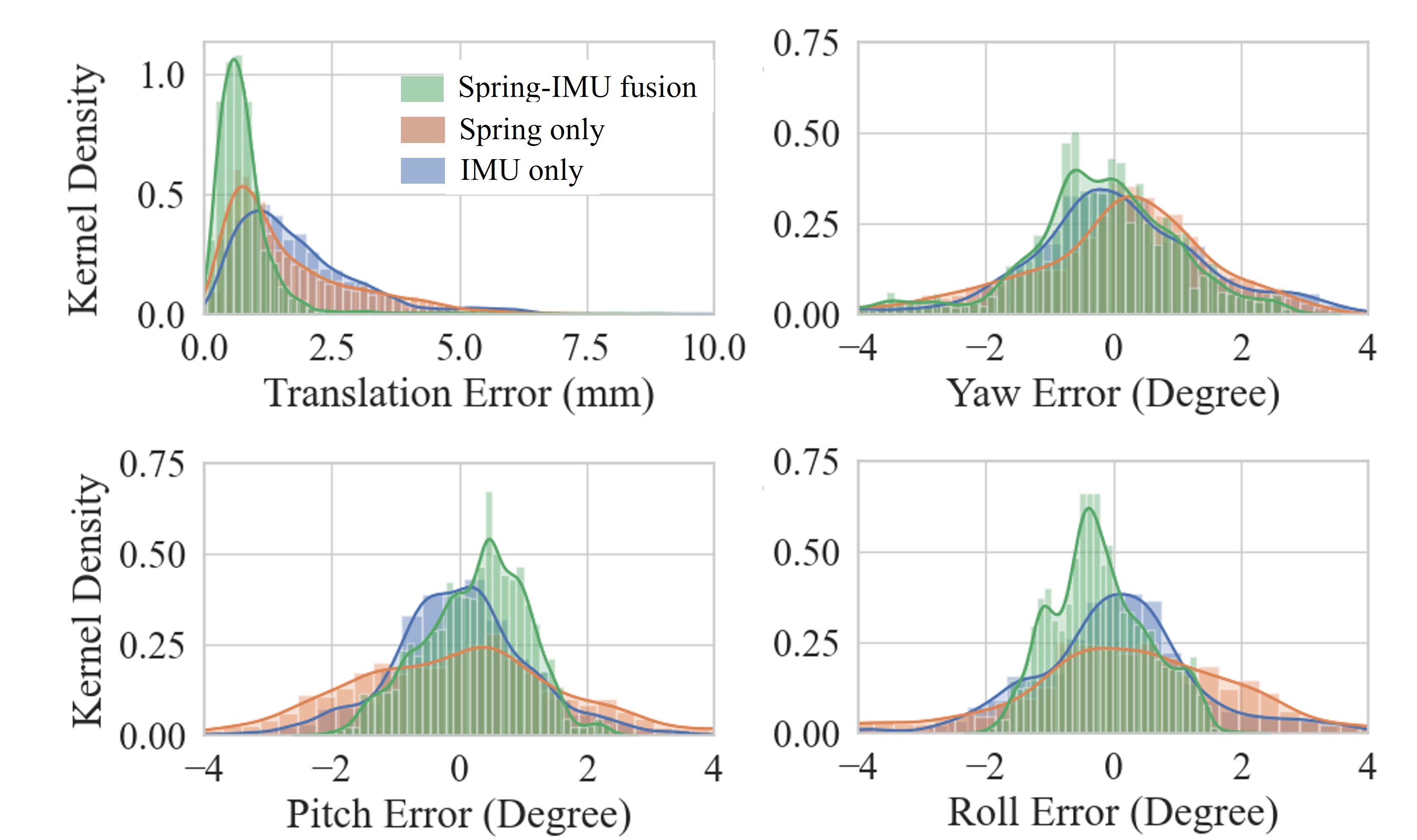} \vspace{2px} \\ 
\footnotesize
\begin{tabular}{l|cccc}
\hline \hline
\specialrule{0em}{2pt}{1pt}
                  & Trans. (mm)$^\dag$ & Yaw $(^\circ)$ & Pitch $(^\circ)$ & Roll $(^\circ)$ \\ 
                  \specialrule{0em}{1pt}{1pt}
                  \hline
                  \specialrule{0em}{1pt}{1pt}
                  		
Spring Only       &   1.64     & 1.33      &  1.53      &   1.67     \\
\specialrule{0em}{1pt}{1pt}
IMU Only          &   1.85     & 1.13      &  0.85      &   1.01     \\
\specialrule{0em}{1pt}{1pt}
Spring-IMU Fusion &   0.90     & 1.06      &  0.71      &   0.66     \\ 
\specialrule{0em}{1pt}{1pt}
\hline \hline
\end{tabular}
\begin{flushleft}
$^\dag$Translation errors are computed by MSE of $\mathbf{T}$ as $\sqrt{\Delta x^2 + \Delta y^2 + \Delta z^2}$
\end{flushleft}
\caption{Accuracy comparison of proprioception when using different sensor signals. Rotational errors of the end-effector (i.e., $\Delta \mathbf{R}$) are decomposed into Yaw, Pitch \& Roll angles for a better illustration. The average errors are also given in the table, where 
$45.1\%$ (translation), $6.19\%$ (Yaw), $16.5\%$ (Pitch), and $34.7\%$ (Roll) 
decreases are observed after applying sensor fusion. 
}\label{fig:sensorComp}
\end{figure}


\subsection{Sensors of Soft Manipulator}\label{subsec:sensorSelection}
The soft manipulator is equipped with a sensor system that is composed of conductive springs and IMUs to realize its proprioception. Specifically, each segment contains six springs placed around the center line of the manipulator and two IMUs installed on the rigid link (see the illustration in Fig.~\ref{fig:setup}(a)). The six conductive springs are used in three pairs, where each pair is placed in a configuration of the `X' shape between two rigid links. 
\revise{We carefully select the type and the length of the springs to avoid the issues of buckling or plastic deformation while large deformation is conducted on the soft manipulator.}
{The extension springs with outer diameter 4.0 mm are made of piano steel wire (with wire diameter as 0.4 mm) to be used in our system to capture inductive signals. A free length of 60 mm is chosen for the springs to avoid buckling during the joint movement. The helix distance is chosen as 0.4 mm, which results in springs with 0.054 N/mm as force rate (i.e., very small forces are generated by their length variation).}
The two IMUs are installed in orientations perpendicular to each other 
to better capture the poses of the rigid pad. To obtain the `ground-truth' of motion for the training and verification purpose, markers for motion capture are installed at the rigid link. Note that the information acquired from a motion capture system will not be used in the control routine.

The conductive springs and IMUs present different advantages to capture the different aspects of the end-effector's poses, where sensor fusion can help integrate their advantages for more accurate proprioception. Ablation study is conducted with results shown in Fig.~\ref{fig:sensorComp}, where the statistics of pose estimation errors by using different sensors are given. It can be found that IMU can provide a relatively good prediction in rotation while springs are better at capturing the translation. When combining the signals from these different sensors together by fusion, the precision accuracy in both translation and rotation can be significantly improved -- see the Table of Fig.~\ref{fig:sensorComp}. 
This result is also reflected by the nature of the working principle of sensors. Springs are more sensitive to the length variation -- i.e., direct measurement of displacements although the length variation on multiple springs can also be converted into orientation change. The IMU signals are obtained by the numerical integration of angular accelerations read acquired from accelerometers, which is more sensitive to the orientation change rather than displacements~\cite{felt2019inductance}. Using the fusion of signals from both the conductive springs and IMUs can take advantage of both sensors and provide more accurate prediction with the help of an effective learning process. Details will be represented in Sec.~\ref{secLearning}. 

\subsection{Processing Sensor Signals and Calibration}
In our system, the signals obtained from the sensor system are collected by Arduino control boards and synchronized by the timestamp method.
The signals from IMU containing Yaw, Pitch, and Roll angles are direct reads 
from the analog input. Kalman filter is used to correct the error with the help of the reads from the IMU's gyroscope and acceleration meters. 


For the spring sensor, we first read its inductance and then convert it to the spring's length change by physical calibration. The electrical circuit presented in Fig.~\ref{fig:setup}(b) can generate a small band of alternating current by continuously charging and discharging a capacitor in parallel to obtain the inductance of the spring. 
A small inductance is connected in series with the spring itself as compensation to adjust the inductance of the spring to an observable range~\cite{sahu2022spring}, which can be computed by
\begin{equation}\label{freq}
  L_{spring} = \frac{1}{C\cdot(2\pi f_{sensor})^2} -L_{comp}.
\end{equation}
Here $C$ is the value of capacitance connected in parallel in the circuit, $f_{sensor}$ is the frequency of the alternating current generated in the circuit, and $L_{comp}$ is the small inductance for value compensation. \revise{}{$L_{comp} = 1000~pF$ is determined via experiment striking a balance between stability and measurement range.}
After reading the inductance value of the spring from the circuit, the spring length is computed via physical calibration as illustrated in the figure above. Data points are generated by using the linear motor to precisely record the elongation of ten springs from their initial length ($70~\mathrm{mm}$) to the maximal length that will be used in the soft manipulator ($240~\mathrm{mm}$). A quartic polynomial function is used to fit the experimental data and obtain a fitting result as
\begin{equation}
    L_{spring}(I)= a + b I + c I^2 + d I^3 + e I^4
\end{equation}
with $a=-12.89$, $b=0.9553$, $c=-1.091 \times 10^{-2}$, $d=5.692 \times 10^{-5}$, $e=-1.119 
\times 10^{-7}$, 
and $I$ being the measured inductance value\revise{.}{~-- see Fig.~\ref{fig:setup}(c).}

We process the sensor signals of springs and IMUs both as geometrically defined features. This is to make a more effective learning process and enable the simulation-based data-efficient learning. Different from the strategy of directly feeding the raw data (i.e., the inductance of springs) to the learning network, the training with geometry-in and geometry-out can effectively reduce the complexity of learning. \revise{}{The resolution of the spring sensor stands at 0.4mm according to our tests.} Moreover, this calibration overcomes the challenge \revise{that the physically-defined inductance is hard to be precisely obtained by a numerical simulator}{of obtaining inductance by simulators} -- i.e., lengths are much easier to \revise{be evaluated.}{measured.}


\section{Learning-based Proprioception with Sensor Fusion and Sim-to-Real Transfer
}~\label{secLearning}
The purpose of learning-based proprioception is to effectively and efficiently construct the mapping function $f_{smap}(\cdot)$ between the signals $\mathbf{s} = [\mathbf{s}_{spring}, \mathbf{s}_{imu}] \in \mathbb{R}^{24}$ and the pose $[\mathbf{T}, \mathbf{R}]$ of the soft manipulator. Here the vector $\mathbf{s}_{spring}$ represents the lengths of twelve springs while 
$\mathbf{s}_{imu} \in \mathbb{R}^{12}$ is compound of the yaw, pitch, and roll angles obtained from the four IMUs. As the coefficients in a rotation matrix are coupled with each other, we convert the $\mathbf{R}$ into yaw, pitch, and roll angles for training. This gives a pose vector $\mathbf{p} = [x,y,z,\theta_{y},\theta_{p},\theta_{r}]\in\mathbb{R}^{6}$ as the output of function $f_{smap}$.
\begin{figure}[t]
\includegraphics[width=\linewidth]{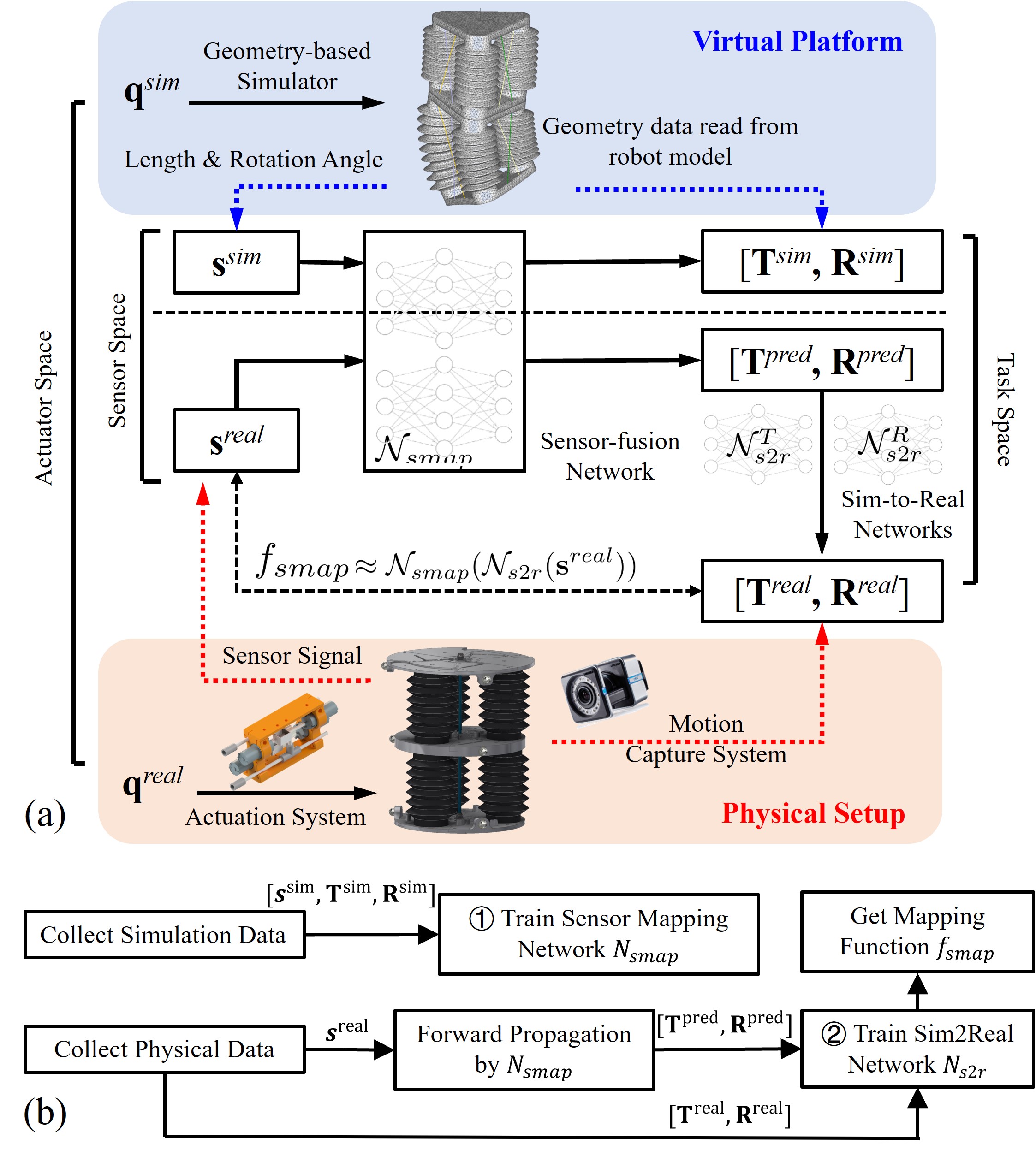}
\caption{(a) The learning pipeline to build the mapping function $f_{smap}$ of proprioception include 1) a \textit{feedforward neural network} (FNN) $\mathcal{N}_{smap}$ trained in simulation and 2) a lightweight network $\mathcal{N}_{s2r}$ for sim-to-real transfer. The networks used in our pipeline are all differentiable, which allows an effective estimation of the Jacobian of control objectives by the backpropagation of the networks. \revise{}{(b) Diagram depicting processes of collecting datasets from both the simulation (a large dataset) and the physical environment (a small dataset), which are used to train $\mathcal{N}_{smap}$ and $\mathcal{N}_{s2r}$ in two steps for constructing the mapping function $f_{smap}$ of proprioception.}
}\label{fig:learningPipeline}
\end{figure}

Our data-efficient learning pipeline is as presented in Fig.~\ref{fig:learningPipeline}. A network $\mathcal{N}_{smap}$ with higher complexity trained by a large dataset generated from the numerical simulation is first used to build the sensor mapping, which is followed by a lightweight sim-to-real network $\mathcal{N}_{s2r}$ to eliminate the error between virtual and physical environments caused by sensor drifting. This sim-to-real transfer design results in a data-efficient training process, where only a small physical dataset is needed to fix the gap between the virtual platform and the physical setup (details presented in Sec.~\ref{subsec:lerningcomparsion}). 

Poses of the soft manipulator in a physical setup can be predicted by the forward propagation of the trained networks as:
\begin{equation}
    [\mathbf{T}^{real}, \mathbf{R}^{real}] \revise{}{= f_{smap}(\mathbf{s}^{real})} 
    \approx 
    \mathcal{N}_{smap}(\mathcal{N}_{s2r}(\mathbf{s}^{real})).
\end{equation}
Such a pipeline of network only using geometry-based features will enable the loading-independent proprioception, which is essential to realize the function of closed-loop control under different external loads. Differently, the previous method proposed in~\cite{fang2022efficient} based on the direct mapping between the actuator space and the task space is loading-dependent. 



\subsection{Network Structure and Data Preparation}\label{subsec:dataGeneration}
\subsubsection{Network for fusion and sensor mapping}\label{sec:Transformer}
We select a \textit{feedforward neural network} (FNN) as the network to achieve sensor fusion and learn the mapping function (denoted by $\mathcal{N}_{smap}$). This helps to establish the nonlinear connection between the sensor space and the task space. The input layer is a flattened vector that contains the signals $\mathbf{s} \in \mathbb{R}^{24}$ from springs and IMUs. The output layer is the pose vector $\mathbf{p}  \in \mathbb{R}^6$. Due to the complexity of signal fusion and sensor mapping, a multi-layer network structure containing two hidden layers with a total of 240 neurons is used to achieve a precise approximation of the mapping function $f_{smap}$. The Tan-Sigmoid activation function is employed in our implementation.

\subsubsection{Network for Sim-to-Real Transfer} After learning the information of sensor fusion and sensor mapping in $\mathcal{N}_{smap}$ (by a dataset obtained from simulation), a lightweight FNN is chosen for the sim-to-real transfer network $\mathcal{N}_{s2r}$. The errors of translation and rotation are trained separately by two networks $\mathcal{N}_{s2r}^{T}$ and $\mathcal{N}_{s2r}^{R}$ to enhance the effectiveness of training as the errors in translation and rotation having significantly different magnitudes. The structure of the network is carefully selected to balance the training accuracy and dataset size, where the over-fitting issue should be avoided (ref.~\cite{fang2022efficient}). In this work, we conduct a network with one hidden layer containing 45 neurons based on experimental tests -- again the Tan-Sigmoid activation function is used.


\subsubsection{Data Generation Process} A geometric-driven simulator~\cite{fang2020kinematics} for soft robots is first used to generate a large dataset to train $\mathcal{N}_{smap}$. Specifically, the actuator space is randomly sampled within the pressure ranges of six chambers to obtain a dataset with $n_{sim} = 20000$ samples. The virtually measured sensor signals (including the lengths of springs and the rotation angles of rigid links) are recorded together with the poses of the end-effector as data points. Due to the efficiency of the simulator, the dataset can be generated in 3 hours and 20 minutes. The process of data generation in the virtual platform has also been demonstrated in the supplementary video.


In the physical setup, a motion capture system containing six Vicon Vero~13 cameras (illustrated in Fig.~\ref{fig:teaser}) is used to collect the ground-truth poses on the end-effector pose (denoted by $[\mathbf{T}^{real},\mathbf{R}^{real}]$). For each recorded physical sensor signals $\mathbf{s}^{real}$, an end-effector's pose can be predicted by $\mathcal{N}_{smap}$ as $[\mathbf{T}^{pred},\mathbf{R}^{pred}] = \mathcal{N}_{smap}(\mathbf{s}^{real})$ which is taken as the input of the sim-to-real network $\mathcal{N}_{s2r}$. The ground-truth pose obtained from the motion capture system is used as the output of $\mathcal{N}_{s2r}$ for training. Based on our tests, a dataset containing 729 samples generated by uniformly sampling each chamber's pressure range (i.e., $n_{real} = 3^6 = 729$) is sufficient enough to capture the sim-to-real `gap', which can be generated within 12 minutes in our experiment.

\begin{figure}[t]
\includegraphics[width=\linewidth]{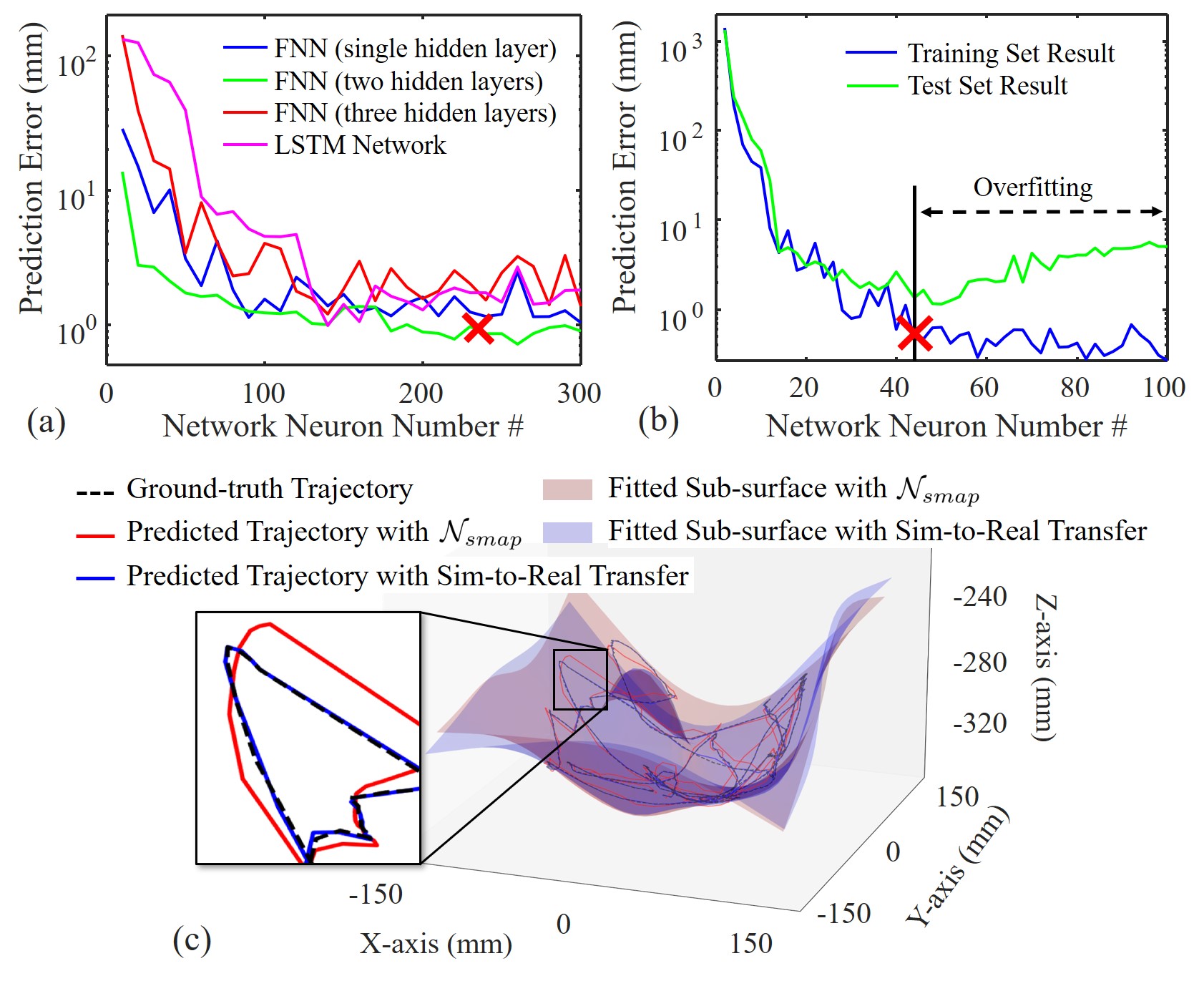}
\caption{Studies about the types of networks, the number of neurons, and the strategy of sim-to-real. 
(a) The study compares the MSEs of the translation predicted by $\mathcal{N}_{smap}$ when using different number of layers in FNNs vs. an LSTM network with the same number of neurons.
(b) The number of neurons in the sim-to-real network $\mathcal{N}_{s2r}$ is carefully studied to avoid over-fitting -- again, it is based on the MSEs of the predicted translation. 
(c) Comparison between the simulation-based prediction (with $20,000$ samples) and the ground-truth for the position of a marker on the end effector -- it can be observed that the gap is small and can be effectively reduced by the sim-to-real transfer (with $729$ samples). Sub-surface embedding of the sampled trajectory points is generated as cubic B-splines by least-square fitting for the purpose of visualization.
}\label{fig:learningResult}
\end{figure}

\vspace{-1 mm}
\subsection{Training Details and Comparison}\label{subsec:lerningcomparsion}
Both networks were implemented by the Matlab machine learning toolbox and trained on a workstation with an NVIDIA 3080 GPU graphics card. We used the mean squared error (MSE) as the loss function. Adam was used as the optimizer for training with a learning rate of $0.01$ and a batch size of $128$. The datasets generated from both the simulator and the physical setup are split into the training, the validation, and the test sets with a ratio of 0.7:0.2:0.1. The training results for both $\mathcal{N}_{smap}$ and $\mathcal{N}_{s2r}$ are presented in Fig.~\ref{fig:learningResult}. It is found that the training process converges effectively on the networks chosen in our work. We also compared the performance of FNN with the result learned on an LSTM network that is time-sequence oriented. However, no significant difference is observed (see Fig.~\ref{fig:learningResult}(a)). This is mainly because the sensor signals in our method are less sensitive to the dynamics of the soft manipulator. More discussions can be found in Sec.\ref{subsec:proprioception}. For the sack of computational efficiency, FNN is chosen in our system. 

We also conducted another study to illustrate the effectiveness of the proposed sim-to-real transfer pipeline. As illustrated in Fig.~\ref{fig:learningResult}(c), the gap between the prediction generated by the simulation according to the measured sensor signals and the ground-truth obtained from the motion capture is relatively small. This is mainly because the sensor signals are represented as geometry-base features and are less influenced by the nonlinearity sourced by soft materials or the actuation system. The observed gap is an integrated error from the damping of the spring sensors and the drifting of IMUs. The gap can be effectively captured by using a data-efficient sim-to-real transfer network using a small dataset with only $729$ samples (see the result with sim-to-real as shown in Fig.~\ref{fig:learningResult}(c)). 

\section{Closed-loop Control via Sensor Space}\label{secControl}
Controlling soft manipulators to perform tasks such as pick-and-place and path following requires the computation of feasible actuation parameters (denoted as $\mathbf{q}$ that represents a point in the actuator space~\cite{george2018control}) that drive the robot to reach a reference target pose $\mathbf{p}^{ref}$. However, existing model-based control strategies based on either analytical models (e.g., PCC~\cite{gerboni2017feedback}) or reduced numerical models (e.g.,~\cite{bieze2018finite}) can become less effective when different large external loads are applied~\cite{azizkhani2022dynamic}. Similarly, the performance of learning-based model-free controllers is also influenced by the variation of external loading. Moreover, the highly nonlinear nature of this problem makes the mapping between the actuator space and the task space a hard learning task to be achieved by using a limited number of samples obtained from physical experiments. Differently, we proposed a control strategy by using information obtained from the sensor space that takes advantage of the precise and robust proprioception on soft manipulators achieved through the geometry-based sensor signals. By using the sensor signals as feedback, our control algorithm is less influenced by the hysteresis of the soft chambers. Moreover, the controller can maintain its functionality when \revise{a variety of large loads are applied to the end-effector of the soft manipulator.}{different weights are applied to the end-effector.} 
\begin{figure}[t]
\includegraphics[width=\linewidth]{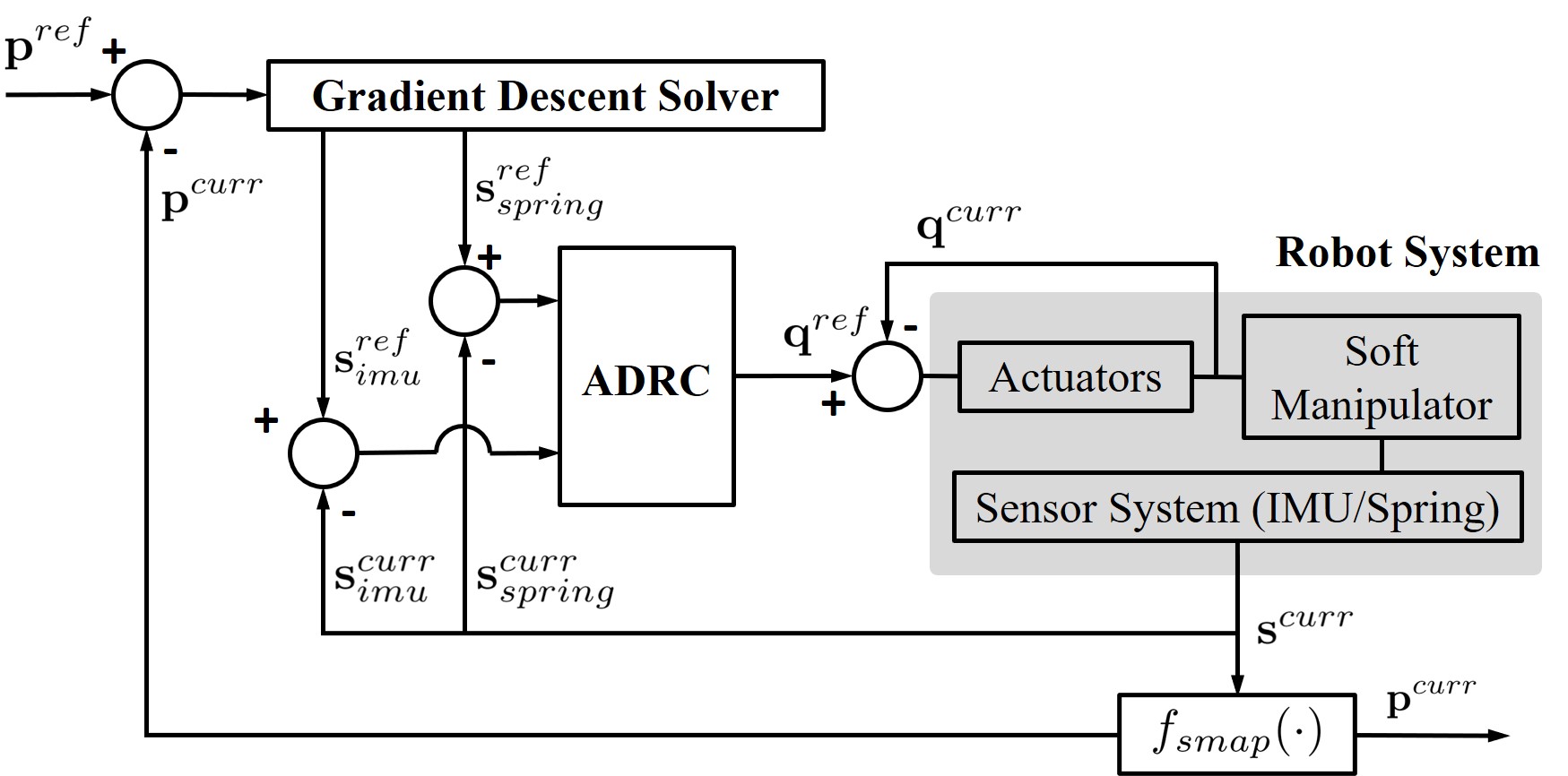}
\caption{An illustration of our sensor-space based control pipeline. A gradient descent solver is invited to iteratively compute reference sensor signals, which is achieved by the ADRC controller. Note that the status of soft manipulator $\mathbf{p}^{curr}$ is evaluated by the learned sensor mapping function $f_{smap}(\cdot)$, which is loading-independent.
}\label{fig:controlPipeline}
\end{figure}

\begin{algorithm}[t]
\label{algGradientDescent}
\caption{Gradient Descent Solver}
\LinesNumbered

\KwIn{Target pose $\mathbf{p}^{ref}$ and the learned network $\mathcal{N}_{smap}$, $\mathcal{N}_{s2r}$.}

\KwOut{Feasible sensor signals $\mathbf{s}^{ref}$.} 

Read initial sensor signal $\mathbf{s}_0$ and compute rest pose $\mathbf{p}_0$;

Set $i=0$, evaluate the objective function $\mathcal{O}(\mathbf{s}_i)$ by Eq.(\ref{eq:controlobjective});

\While{$\mathcal{O}(\mathbf{s}_i) > \lambda$ and $i<i_{max}$}{

Evaluate $d\mathcal{O}/d\mathbf{s}_i = -2(\mathbf{p}^{ref} - f_{smap}(\mathbf{s}_i))\cdot \mathcal{N}_{smap}^{-1} \mathcal{N}_{s2r}^{-1} $;

Compute $\mathcal{O}_{shrink}=\mathcal{O}(\mathbf{s}_i + h \frac{d\mathcal{O}}{d\mathbf{s}_i})$;

\tcc{Line search to find optimal step $h$.}

\While{$\mathcal{O}_{shrink} > \mathcal{O}(\mathbf{s}_i)$~~\tcc{Step 1: Shrinking}}{ 
$h=\tau h$, compute and update $\mathcal{O}_{shrink}=\mathcal{O}(\mathbf{s}_i + h \frac{d\mathcal{O}}{d\mathbf{s}_i})$;
}

Set $\mathcal{O}_{expand} = \mathcal{O}_{shrink}$; \tcc{Step 2: Expanding}
\Repeat{$\mathcal{O}_{new} < \mathcal{O}_{expand}$}{

Set $\mathcal{O}_{new} = \mathcal{O}_{expand}$ and $\mathbf{s}_i = \mathbf{s}_i + h$;

Compute and update $\mathcal{O}_{new}=\mathcal{O}(\mathbf{s}_i + h \frac{d\mathcal{O}}{d\mathbf{s}_i})$;

}

$i = i+1$, set $\mathcal{O}(\mathbf{s}_i) = \mathcal{O}_{new}$;

}

\textbf{return} $\mathbf{s}^{ref} = \mathbf{s}_i$;

\end{algorithm}


\subsection{Sensor-Space Based Control Algorithm} \label{subsec:controller}

The proposed control framework is as shown in Fig.~\ref{fig:controlPipeline} which includes a few modules that run iteratively until the system converges and the soft manipulator reaches target pose $\mathbf{p}^{ref}$. Our closed-loop control architecture is formed by 
\begin{itemize}
    \item A \textit{gradient descent} (GD) solver which computes a feasible sensor signal $\mathbf{s}^{ref}$ based on the given target $\mathbf{p}^{ref}$ with the help of differentiable $\mathcal{N}_{smap}$ and $\mathcal{N}_{s2r}$;
    \item An \textit{active disturbance rejection controller} (ADRC) \cite{han2009pid} to generate a reference actuation parameter $\mathbf{q}^{ref}$ that can drive the robot system to achieve $\mathbf{s}^{ref}$.
\end{itemize}
Details are presented below.

\begin{figure*}[t]
\includegraphics[width=\linewidth]{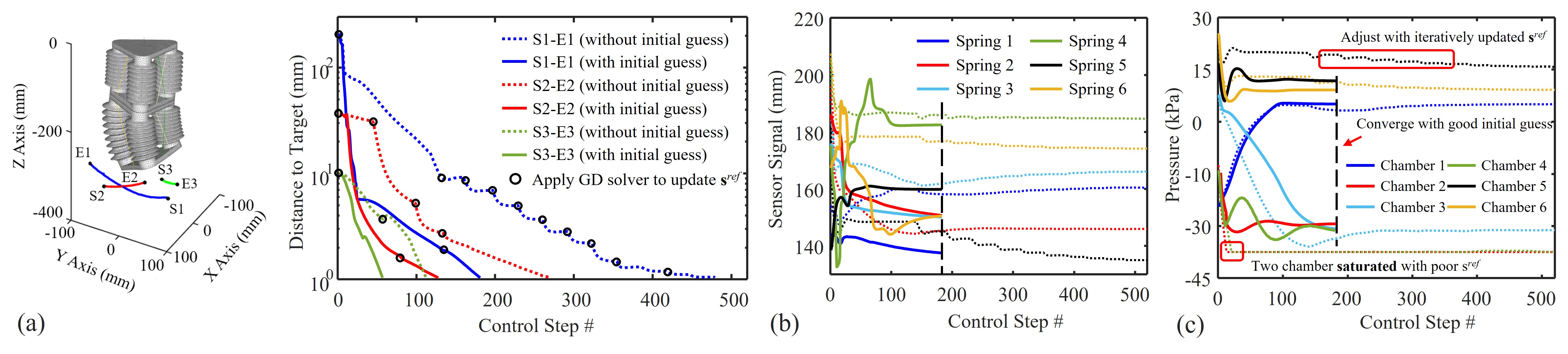}
\caption{Performance verification of the proposed control algorithm running on a virtual platform with three different target poses to achieve. (a) Visualization of the approaching trajectories and the convergence of computation throughout the control steps in the \textit{task spaces}. 
(b) The change of signals of in the \textit{sensor space} for approaching Target 1. Solid lines and dash lines are employed to respectively denote the results with and without a good initial guess of $\mathbf{s}^{ref}$. 
(c) The change of pressures in the \textit{actuator space} to iteratively achieve the target position 1. It can be found that a poor initial guess may lead to the saturation of chambers as the pressure reaches its limits (i.e., the situation presented by dash lines).
}\label{fig:controlVirtual}
\end{figure*}

With a given reference pose $\mathbf{p}^{ref}$, the gradient descent solver presented in \textbf{Algorithm~I} is used to computes a feasible sensor signal $\mathbf{s}^{ref} = [\mathbf{s}^{ref}_{imu}, \mathbf{s}^{ref}_{spring}]$ that minimize the control objective defined as 
\begin{equation}\label{eq:controlobjective}
    \mathcal{O}_{outer} = \| \mathbf{p}^{ref} - \mathbf{p}^{curr}  \|^2 \approx \| \mathbf{p}^{ref} - f_{smap}(\mathbf{s}^{curr})  \|^2.
\end{equation}
We apply the sensor mapping function $f_{smap}$ as cascaded networks $\mathcal{N}_{smap}$ and $\mathcal{N}_{s2r}$ learned in the previous section to precisely predict the current pose of the soft manipulator by the signals $\mathbf{s}^{curr}$ read from the sensor system. Since $f_{smap}$ is independent to the external loads, the computed $\mathbf{s}^{ref}$ can always tend to drive the soft manipulator toward the target pose $\mathbf{p}^{ref}$ regardless of loading conditions. In our implementation, the gradient of the objective $d\mathcal{O}/d\mathbf{s}$ is evaluated by chain rules with the differentiation of networks (i.e., denoted by~$\mathcal{N}^{-1}_{smap}$ and $\mathcal{N}^{-1}_{s2r}$). A line search method with shrinking ratio $\tau = 0.2$ (see also line $6-14$ of the \textbf{Algorithm~I}) is used to guarantee the convergence of our algorithm where the value of the objective function will be ensured to decrease in every iteration~\cite{fang2020kinematics}. Thanks to the high efficiency of computation on both the forward and the backward propagation of the networks, the average time to find a feasible sensor signal is about $17~\mathrm{ms}$ per target point. This efficient GD solver runs as the outer loop of the control algorithm to continuously update $\mathbf{s}^{ref}$ based on the soft manipulator's status $\mathbf{p}^{curr}$.

The computed $\mathbf{s}^{ref}$ then serves as the target control goal in the inner loop of our control algorithm, where the ADRC controller is employed to adjust the pressures in all chambers according to the difference between $\mathbf{s}^{curr}$ and $\mathbf{s}^{ref}$. The output ADRC controller is a reference actuation parameter $\mathbf{q}^{ref}$ to be realized by the actuation system meanwhile considering the current actuation parameter $\mathbf{q}^{curr}$. Note that the sensor system used in our setup invites redundancy (i.e., the dimension of sensor space is higher than the dimension of the task space), controlling $\mathbf{q}$ to reach $\mathbf{s}^{ref}$ is an under-determined problem. There is no guarantee that a reference actuation parameter $\mathbf{p}^{ref}$ can always be found to reach $\mathbf{s}^{curr}$ in the sensor space exactly. Therefore, the ADRC controller in fact tends to minimize $\mathcal{O}_{inner} = \| \mathbf{s}^{curr} -\mathbf{s}^{ref} \|^2$. When $\mathcal{O}_{inner}$ is converged into a local minimum, the value of $\mathbf{s}^{ref}$ will be updated by the GD solver in the outer loop with reference to a newly achieved pose being estimated as $\mathbf{p}^{curr}$ through $f_{smap}(\cdot)$. These two loops are computed iteratively until the system converges (i.e., $\mathcal{O}_{outer}$ is minimized). 
In the following sub-section, we will study the effectiveness of our control algorithm on virtual platforms. Its performance of controlling soft manipulator in physical tests for the tasks of pick-and-place and trajectory following will be discussed in Sec.~\ref{subsec:pickandplaceControl}. 

\subsection{Performance Verification and Selection of Initial Guess}
The performance of our sensor-space based control algorithm is verified on a virtual platform for point-to-point movements with trajectories of varying distances. The performance in task spaces is as shown in Fig.~\ref{fig:controlVirtual}(a), where our control algorithm can successfully achieve the tasks. We found that the convergence of the proposed control method is also highly influenced by the distance between the initial pose $\mathbf{p}^{curr}$ and the target $\mathbf{p}^{ref}$. It can be found from the visualization of the sensor signal shown in Fig.~\ref{fig:controlVirtual}(b) that the longest path (trajectory 1) contains some periods where the reachability of computed $\mathbf{s}^{ref}$ cannot be realized due to the chamber saturation (i.e., chambers 2 and 4 demonstrated in Fig.~\ref{fig:controlVirtual}(c)). Therefore, the proposed controller takes more iteration steps to update $\mathbf{s}^{ref}$ and converge to the final destination.  

To accelerate the convergence of the control algorithm, we developed a method for selecting a good initial guess for the first GD-solving step. Specifically, a feasible sensor signal $\mathbf{s}_0$ is searched among the dataset collected to train $\mathcal{N}_{smap}$. This initial guess of sensor singles should minimize $\mathcal{O}_{outer}$ meanwhile avoid the chambers being saturated (i.e., the determined pressures are within the min/max allowed ranges). This strategy can highly improve the feasibility of $\mathbf{s}^{ref}$ computed by GD solver and meanwhile resulting in up to three times acceleration in the control process (i.e., 160 vs. 480 iterations) for the long-distance case. Our iterative-based control algorithm can converge in a few control steps for all three tested target poses. 

\section{\revise{Result and Discussion}{Experimental Results}}\label{secResult}
The proposed method of spring-IMU fusion based proprioception and the algorithm for closed-loop control have been implemented and tested on the soft manipulator as shown in Fig.~\ref{fig:teaser}. We conduct all the computation on a laptop with an Intel 12700H CPU (14 Cores @2.3GHz) and 32 GB of memory, where the trained networks used to predict the pose and control the soft manipulator are integrated into a C++ program based on QT framework. Meanwhile, OpenMP library is used to support parallel computing to enable the highest computational efficiency when evaluating the sensor-pose mapping function $f_{smap}$. Note that for all the physical experiments presented in this section, the Vicon motion capture system is only used to collect ground-truth poses \revise{of the soft manipulator for verifying the performance of proprioception and control}{for verification}. The vision information was not employed to provide feedback for control. 

\begin{figure}[t]
\centering
\includegraphics[width=\linewidth]{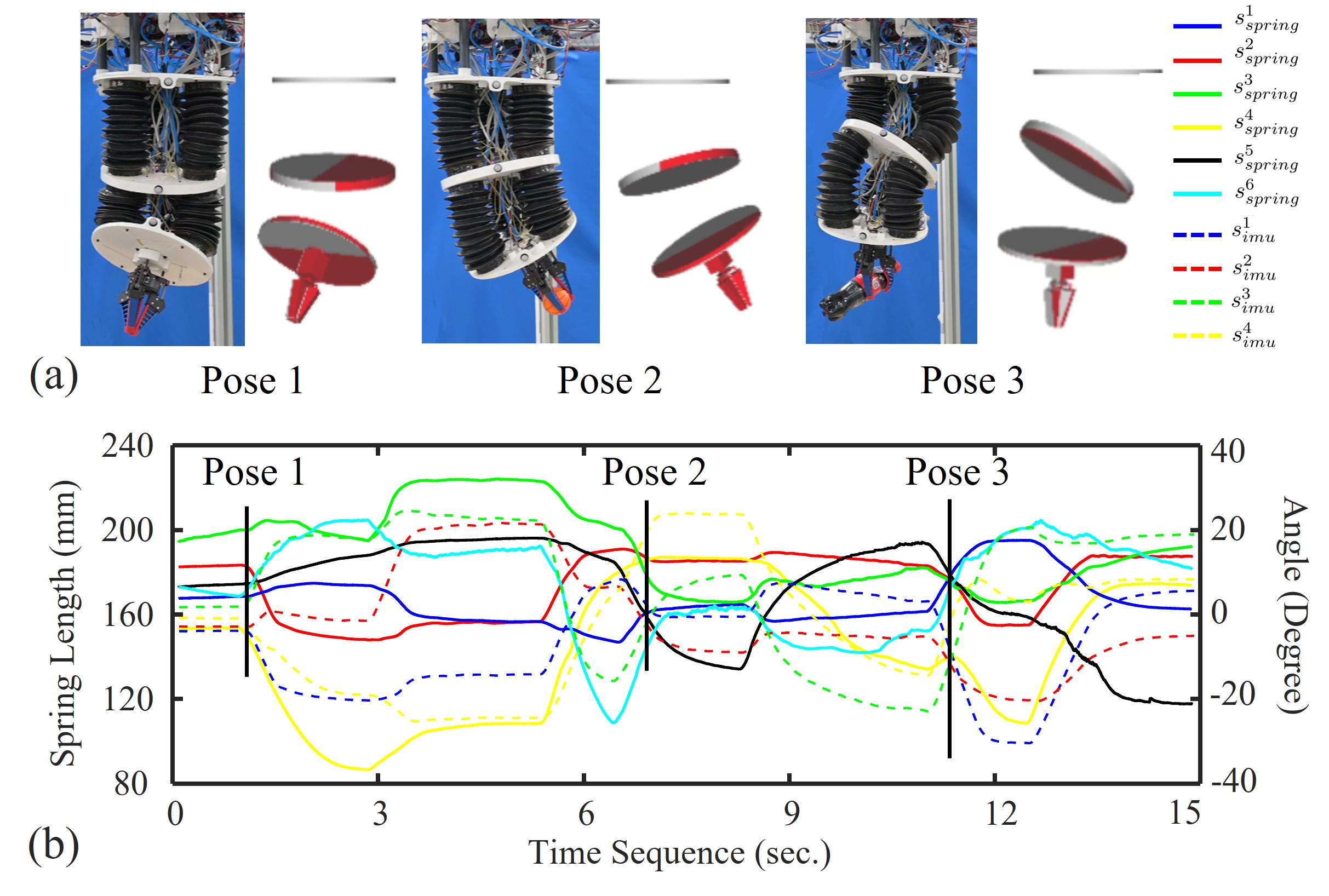}
\caption{(a) Comparisons of the predicted poses (shown in red) on soft manipulators with the ground-truth poses (shown in gray). (b) corresponding spring and IMU signals when different external loads are applied at the end-effector of the soft manipulator. 
}\label{fig:3DProperception}
\end{figure}

\subsection{Proprioception with External Loading}\label{subsec:proprioception}
We first tested fusion-based the proprioception of soft manipulators. The performance is tested by applying different external loads varied from 0g to 500g ($60\%$ of the soft manipulator's weight). Specifically, we evaluated the accuracy of our method in predicting the Yaw, Pitch \& Roll angles and the positions of the end-effector. Comparisons of the predicted poses v.s. the ground-truth poses have been visualized in Fig.~\ref{fig:3DProperception}(a), where different external loads were applied. The predicted poses of the rigid link match well with the ground-truth collected by the motion capture system. The corresponding sensor signals of three poses with different external loads are also visualized in Fig.~\ref{fig:3DProperception}(b). The performance of pose prediction has also been demonstrated in the supplemental video attached with the manuscript and can also be found at \url{https://youtu.be/IqMtvoxj5Jc}.


\begin{figure}[t]
\includegraphics[width=\linewidth]{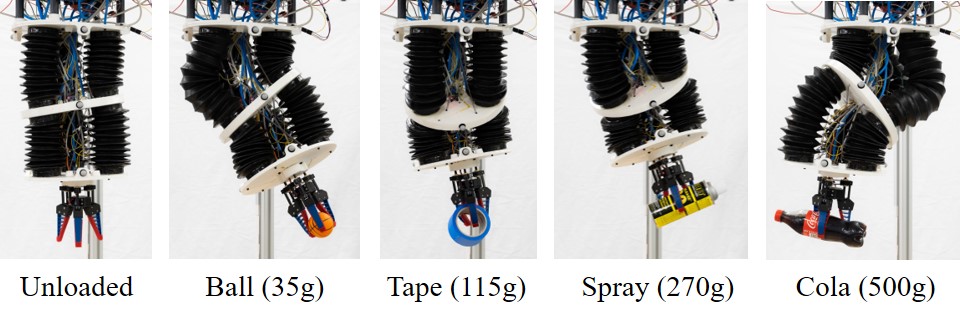}
\\
\footnotesize
\begin{tabular}{l|cccc}
\hline \hline
\specialrule{0em}{2pt}{1pt}
                  & Translation (mm) & Yaw $(^\circ)$ & Pitch $(^\circ)$ & Roll $(^\circ)$ \\ 
                  \specialrule{0em}{1pt}{1pt}
                  \hline
                  \specialrule{0em}{1pt}{1pt}
Unloaded       &   0.90   &   1.06    &   0.71    &   0.66      \\
\specialrule{0em}{1pt}{1pt}
Ball (35g)     &   0.89   &   1.07    &   0.69    &   0.84      \\
\specialrule{0em}{1pt}{1pt}
Tape (115g)    &   0.93   &   1.21    &   0.68    &   0.55      \\
\specialrule{0em}{1pt}{1pt}
Spray (270g)   &   0.83   &   1.03    &   0.40    &   0.43      \\
\specialrule{0em}{1pt}{1pt}
Cola (500g)    &   0.94   &   1.69    &   0.47    &   0.48      \\
\specialrule{0em}{1pt}{1pt}
\hline \hline
\end{tabular}
\caption{Quantitative analysis of the prediction errors (i.e., MSE) in translation and rotation when applying different external loads.}\label{fig:loadingCompare}
\end{figure}

The quantitative analysis of the prediction errors is given in the table of Fig.~\ref{fig:loadingCompare}, where the sensor mapping network is trained using massive simulation data and limited physical data without applying any external load on the end-effector (as presented in Sec.~\ref{subsec:dataGeneration}). It can be found that the errors remain consistently small across all tested cases with an average error of less than 1.0 mm in translation and 1.2 degrees in rotation under a variety of external loads. 
This result demonstrates that the proposed sensing method and the learning-based mapping are robust to different loading conditions. In short, the accuracy in predicting the manipulator's poses is less influenced by the change of external loads. 

\begin{figure}[t]
\includegraphics[width=\linewidth]{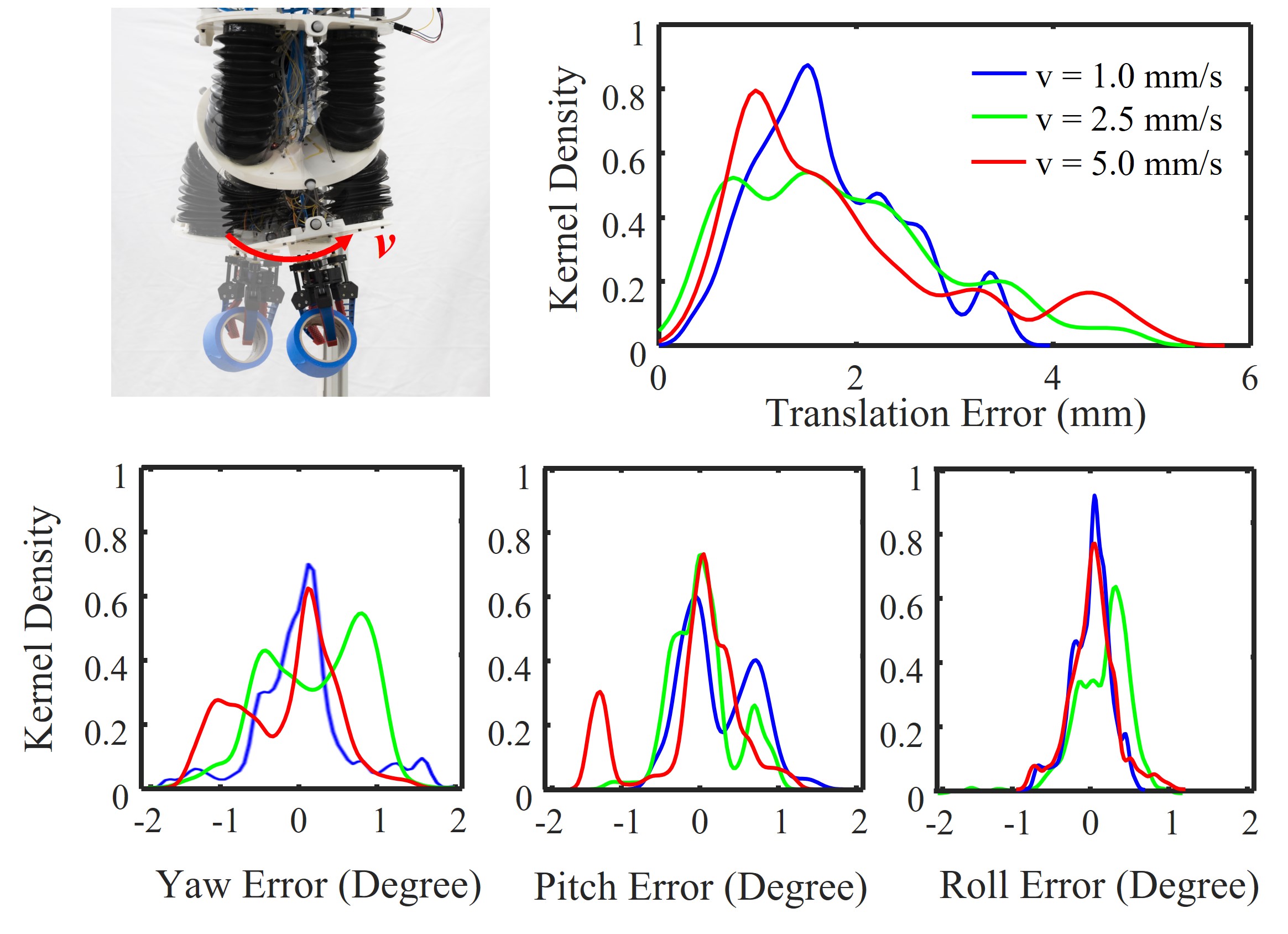}
\\
\footnotesize
\begin{tabular}{l|cccc}
\hline \hline
\specialrule{0em}{2pt}{1pt}
                  & Translation (mm) & Yaw $(^\circ)$ & Pitch $(^\circ)$ & Roll $(^\circ)$ \\ 
                  \specialrule{0em}{1pt}{1pt}
                  \hline
                  \specialrule{0em}{1pt}{1pt}
$1.0~mm/s$       &   0.96   & 0.99      &   0.78    &   0.44      \\
\specialrule{0em}{1pt}{1pt}
$2.5~mm/s$     &   0.88   &  1.27     &   0.77    &   0.58      \\
\specialrule{0em}{1pt}{1pt}
$5.0~mm/s$    &   0.92   &   1.03    &   0.93    &    0.47     \\
\specialrule{0em}{1pt}{1pt}
\hline \hline
\end{tabular}
\caption{Comparison of the proprioception accuracy when applying different speeds of motion on the end-effector of the soft manipulator. The accuracy is less influenced by the speed of motion due to the robustness of our geometry-based sensing for proprioception.}\label{fig:speedTest}
\end{figure}
In practical applications, soft manipulators can exhibit highly dynamic behavior when undergoing deformation. To test the robustness of our proprioception method in such a scenario, experiments have been taken to evaluate the performance of our system under different actuation speeds. Specifically, we applied actuation speeds ranging from 1.0 mm/s to 5.0 mm/s (the maximum allowed speed by our actuation system) and found that the accuracy of pose sensing was not significantly influenced even after applying the highest speed of actuation (as shown in Fig.~\ref{fig:speedTest}). Our system can provide accurate feedback across a wide range of operations, with the predicted translation error remaining at $0.7\%$ across the workspace (size: $300 \times 300 \times 300$ mm).

\begin{figure*}[t]
\includegraphics[width=\linewidth]{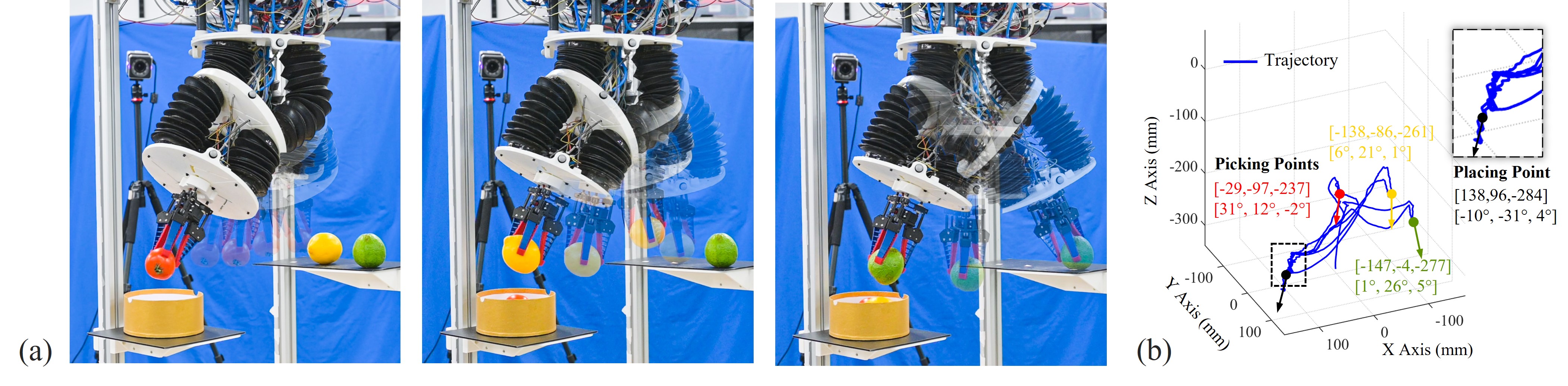}
\caption{(a) With the help of sensor signals to conduct closed-loop control, the soft manipulator can successfully conduct the pick-and-place task for fruits in different weights. (b) The trajectories are visualized in the task space.
}\label{fig:pickandplace}
\end{figure*}

\subsection{Sensor-based Closed-loop Control for Pick-and-Place}~\label{subsec:pickandplaceControl}
Based on the precise proprioception, experiments have been conducted to test the performance of our sensor-space based control algorithm for the soft manipulator. The system was first tested on the pick-and-place tasks by controlling the manipulator's end-effector to reach a set of predefined target poses. As illustrated in Fig.~\ref{fig:pickandplace}(a), the soft manipulator is controlled to reach the target picking point. Then, the gripper can successfully tightly grasp fruits with different weights, including tomato (120 g), lemon (210 g), and guava (170 g). The proposed control algorithm can drive the soft manipulator to the placing point and put the object in the basket. The trajectories have been visualized in the task space (see Fig.~\ref{fig:pickandplace}(b)), where we found that the proposed sensor-space based control can provide smooth movement between two points with large pose variations. The motion was detected to become less stable near the placing point (see the zoom-in figure of Fig.~\ref{fig:pickandplace}(b)). This is because our control algorithm needs to iteratively converge to the final stable status. 
The performance of the pick-and-place process can also be seen in the supplementary video of this paper.


\begin{figure}[t]
\includegraphics[width=\linewidth]{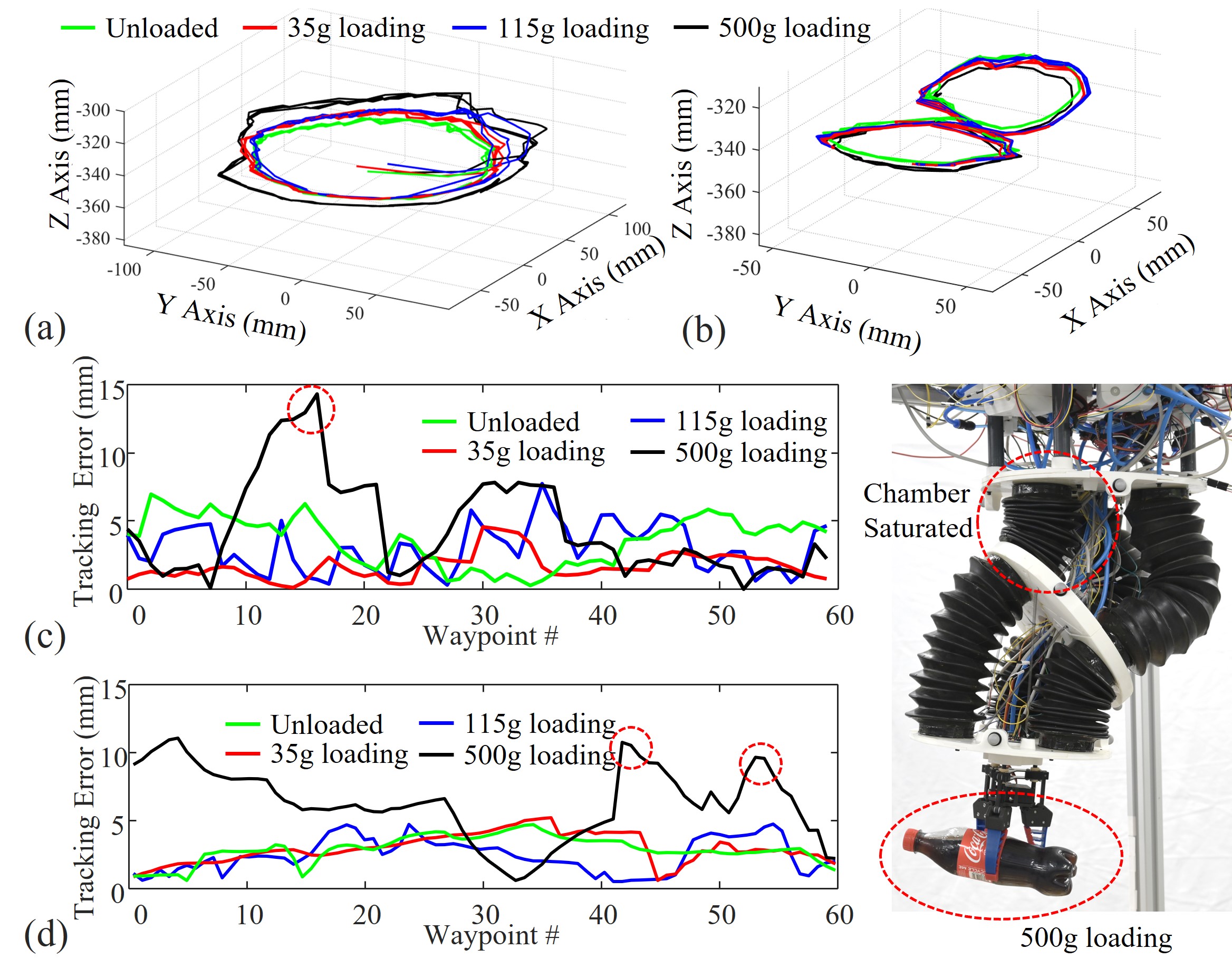}
\caption{Results of path following with different external loads for (a) a circular path and (b) an 8-shape path. Tracking errors are shown for (c) the circular and (d) the 8-shape paths respectively.}\label{fig:trajectory}
\end{figure}

\subsection{Closed-loop Control for Path Following}~\label{subsec:trajFollow}
We further evaluate the performance of our sensor-based approach for the path following tasks by the soft manipulator. Specifically, we control the center of the rigid pad on the end effector to follow predefined paths represented by a set of waypoints as the target pose. In particular, we tested the tasks on two different paths including a circle and an 8-shape. Both of them contain 60 waypoints that are uniformly distributed in the task space. The manipulator is controlled to reach waypoints by using our sensor-space based control algorithm. For these path following tasks, we conduct the converged sensor signals for the previous waypoint as the initial guess of $\mathbf{s}^{ref}$ used in the GD solver (presented in Sec.~\ref{subsec:controller}) for the new waypoint. This leads to very fast convergence in computation as these waypoints are very close to each other. 

The results of the path following tests are shown in Fig.~\ref{fig:trajectory}. It can be easily found that our control algorithm can drive the soft manipulator to accurately track the predefined path. The average tracking errors are 3.7~mm and 3.3~mm for the circle and the 8-shape paths respectively. These tracking errors are within 5.6\% of the workspace dimension when the external load is less than 200~g. On the other hand, it is not surprising to see that the tracking error in a certain pose becomes much larger when a heavier object (e.g., the 500~g cola bottle) is loaded -- as highlighted by red dash circles in Fig.~\ref{fig:trajectory}(c) and Fig.~\ref{fig:trajectory}(d). This is due to the saturation of the chamber when a large external load is applied, which reduces the workspace and makes the computed reference sensor signal hard to be realized The situation is similar to what we already showed on the virtual platform in Fig.~\ref{fig:controlVirtual}(b). 

\section{Discussion}\label{secDiscussion}
%
\revise{}{In this section, we discuss a few limitations of our work, including the proposed sensing system, the controller and the setup of soft manipulator.}

\subsection{\revise{}{Performance of IMU-spring Based Proprioception}}
Experimental results demonstrate that the spring-IMU fusion proposed in our work can realize accurate and robust proprioception on the soft manipulator with a large workspace. The sensor signals used for learning are geometry based, which is loading-independent and also enables the use of geometry-oriented simulator in the sim-to-real pipeline. 
In contrast to existing methods such as the strain gauge approach presented in~\cite{zhao2021shape} that suffers from reduced accuracy in sensing when different external loads are applied, our method has demonstrated its strong capability and 
robustness in physical experiments. \revise{}{Compare with the method presented in our prior work using light-dependent resistors (LDR) sensors~\cite{3D-Deformation-sensing}, the fusion of spring and IMU sensors provides much more stable signals even with tremendous joint bending up to $80^{\circ}$. The capability of load-independent and stable sensing in large deformation allows the proposed proprioception method to become} promising for real-world applications where soft manipulators are operated in different scenarios and interact with objects of varying sizes and weights. 

\begin{figure}[t]
\includegraphics[width=\linewidth]{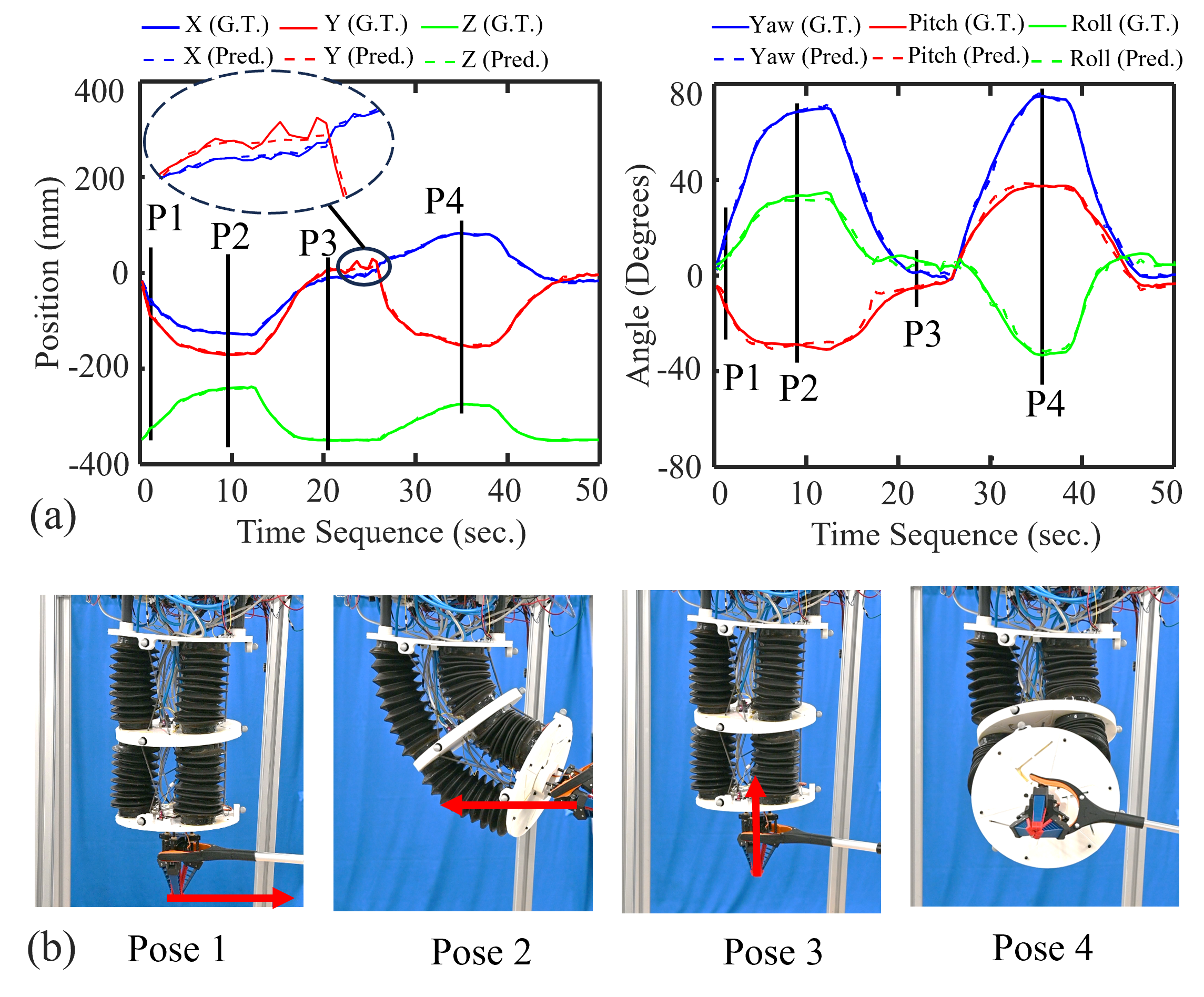}
\\
\footnotesize
\resizebox{88mm}{6mm}{ 
\begin{tabular}{l|cccccc}
\hline \hline
\specialrule{0em}{1pt}{1pt}
                   & Yaw $(^\circ)$ & Pitch $(^\circ)$ & Roll $(^\circ)$ & X (mm)  & Y (mm)  & Z (mm) \\ 
                  \specialrule{0em}{1pt}{1pt}
                  \hline
                  \specialrule{0em}{1pt}{1pt}
$MSE$       &    1.36   & 1.25      & 1.49    & 1.49  &   1.47 &   0.91  \\
\specialrule{0em}{1pt}{1pt}

\hline \hline
\end{tabular}}
\caption{\revise{}{The proposed IMU-spring sensor system can provide precise propriociption when applying lateral forces to the soft manipulator. (a) The predicted position and orientation of end-effector using sensor signal matches well with the ground truth (G.T.). (b) Corresponding poses of the manipulator under different lateral forces. 
}\label{fig:lateralForce}}
\end{figure}

\revise{Future work can focus on optimizing the locations of sensors and the number of sensors. Additionally, applying the proposed sensor system to other soft manipulator systems could  further explore its functionality in other applications.}{
We also tested the performance of the proposed sensing method in a broader scenarios of external loading. As shown in Fig.~\ref{fig:lateralForce}, lateral forces of $40~N$ (measured by force sensor) are applied to the end-effector from different directions using a rod pickup tool. Precise proprioception is maintained in these tests. However,
we also find a limitation in our method when applying to sudden shocks or disturbances. Accuracy in predicting the translation at end-effector is inevitably influenced (see the zoom-in view in Fig.~\ref{fig:lateralForce}(a)). This is due to the damping of the spring sensor, which introduces a certain level of delay in the signal by the disturbances. We plan to enlarge the stiffness of springs in the future to reduce these vibrations.}


\subsection{\revise{}{Limitation on Sensor-Space Based Controller}}
\revise{}{Effectiveness of the proposed closed-loop controller has been validated through physical experiments.} 
\revise{Our proposed sensor-space-based control pipeline has demonstrated promising performance for the control of soft manipulators using sensor signals as feedback. }{It outperforms the learning-based kinematic controller presented in our previous research~\cite{fang2022efficient}. When external loading is applied, the controller without sensor feedback~\cite{fang2022efficient} can easily become ineffective. Differently, the sensor-space-based controller consistently maintains its functionality for different loads with up to 500g. This enhances the manipulator's effectiveness in practical applications like pick-and-place heavy objects as already demonstrated in Fig.~\ref{fig:pickandplace}.} However, as mentioned earlier, the redundancy in sensor space presents a challenge when the target pose is far from the current status, leading to slow convergence in control due to chamber saturation.
Although we proposed a method to select a good initial guess for the \revise{iterative-based}{iteration of} GD solver, more effective techniques such as null space control~\cite{chu2022feedback} can be applied to ensure a feasible \revise{target sensor signal being computed. By optimizing the control objectives within the null space of the soft manipulator's joint space, the system can effectively}{solution while} \revise{avoid}{avoiding} singularities and \revise{reduce}{reducing} the risk of chamber saturation. However, this requires knowledge of the relationship between the actuator and the sensor spaces, which is challenging especially when uncertain large loads are applied to the system. One possible solution for future work is to \revise{employ}{seek the help of} adaptive learning (e.g.,~\cite{Model-based-sensing}). \revise{We plan to investigate this method in our future work.}{}

\subsection{\revise{}{Limitation on Soft Manipulator and Its Setup}}
\revise{}{
Our soft manipulator is currently designed by using relatively soft bellows so that low pressure pump is employed to actuate the deformation to maximize the safety while interacting with humans and the environment. However, the low-pressure actuation restricts the manipulator's workspace and its payload capacity. Moreover, it also limits the capability of operating the manipulator in different setup orientations (e.g., being positioned horizontally or laid on the ground), primarily due to less ability to resist gravity. We plan to improve the chamber's rigidity and employ higher pressures for actuation in our future design. Specifically, we plan to keep the highest pressure within one bar, which is in line with safety and mobility considerations as suggested in~\cite{zughaibi2021fast, li2022equivalent}. This will also help to realize a soft manipulator with more than two segments while enhancing the capability to handle collision and interruptance in our future work.
}

\section{Conclusion}\label{secConclusion}
In this work, we present a novel framework to realize proprioception and closed-loop control for soft manipulators under large deformation. Our approach uses a fusion-aware learning pipeline that combines conductive spring sensors and IMUs to effectively establish the mapping from sensor signals to the poses of the soft manipulator. Multiple geometric signals are fused into robust pose estimations, and a data-efficient training process is achieved after applying the strategy of sim-to-real transfer. We have demonstrated the effectiveness of our approach through physical experiments. The result achieves load-independent and accurate proprioception with an average prediction error under $1~mm$ in translation and $1.2$~degrees in rotation.
The realized proprioception on soft manipulator is thereafter contributed to building a sensor-space based algorithm for closed-loop control. The sensor-space based control algorithm proposed in our work converges very fast and shows strong robustness, which allows the soft manipulator to complete tasks such as path following and pick-and-place under different external loads. Our approach demonstrated a promising framework for sensing and control of soft manipulators with complex and large deformation. \revise{Future work can be investigated to apply this methodology to the applications with more complex tasks on other soft robotic systems.}{}


\bibliographystyle{IEEEtran}
\bibliography{TMECHSoroSensing}

\begin{thebibliography}{10}
\providecommand{\url}[1]{#1}
\csname url@rmstyle\endcsname
\providecommand{\newblock}{\relax}
\providecommand{\bibinfo}[2]{#2}
\providecommand\BIBentrySTDinterwordspacing{\spaceskip=0pt\relax}
\providecommand\BIBentryALTinterwordstretchfactor{4}
\providecommand\BIBentryALTinterwordspacing{\spaceskip=\fontdimen2\font plus
\BIBentryALTinterwordstretchfactor\fontdimen3\font minus \fontdimen4\font\relax}
\providecommand\BIBforeignlanguage[2]{{%
\expandafter\ifx\csname l@#1\endcsname\relax
\typeout{** WARNING: IEEEtran.bst: No hyphenation pattern has been}%
\typeout{** loaded for the language `#1'. Using the pattern for}%
\typeout{** the default language instead.}%
\else
\language=\csname l@#1\endcsname
\fi
#2}}

\bibitem{falkenhahn2016dynamic}
V.~Falkenhahn, A.~Hildebrandt, R.~Neumann, and O.~Sawodny, ``Dynamic control of the bionic handling assistant,'' \emph{IEEE/ASME Transactions on Mechatronics}, vol.~22, no.~1, pp. 6--17, 2016.

\bibitem{hughes2016soft}
J.~Hughes, U.~Culha, F.~Giardina, F.~Guenther, A.~Rosendo, and F.~Iida, ``Soft manipulators and grippers: a review,'' \emph{Frontiers in Robotics and AI}, vol.~3, p.~69, 2016.

\bibitem{ranzani2015bioinspired}
T.~Ranzani, G.~Gerboni, M.~Cianchetti, and A.~Menciassi, ``A bioinspired soft manipulator for minimally invasive surgery,'' \emph{Bioinspiration \& biomimetics}, vol.~10, no.~3, p. 035008, 2015.

\bibitem{gong2019opposite}
Z.~Gong, B.~Chen, J.~Liu, X.~Fang, Z.~Liu, T.~Wang, and L.~Wen, ``An opposite-bending-and-extension soft robotic manipulator for delicate grasping in shallow water,'' \emph{Frontiers in Robotics and AI}, vol.~6, p.~26, 2019.

\bibitem{Nature2015}
D.~Rus and M.~Tolley, ``Design, fabrication and control of soft robots,'' \emph{Nature}, vol. 521, no. 7553, pp. 467--475, 2015.

\bibitem{della2020model}
C.~Della~Santina, R.~K. Katzschmann, A.~Bicchi, and D.~Rus, ``Model-based dynamic feedback control of a planar soft robot: trajectory tracking and interaction with the environment,'' \emph{The International Journal of Robotics Research}, vol.~39, no.~4, pp. 490--513, 2020.

\bibitem{huang2021kinematic}
X.~Huang, J.~Zou, and G.~Gu, ``Kinematic modeling and control of variable curvature soft continuum robots,'' \emph{IEEE/ASME Transactions on Mechatronics}, vol.~26, no.~6, pp. 3175--3185, 2021.

\bibitem{fang2020kinematics}
G.~Fang, C.-D. Matte, R.~B. Scharff, T.-H. Kwok, and C.~C. Wang, ``Kinematics of soft robots by geometric computing,'' \emph{IEEE Transactions on Robotics}, vol.~36, no.~4, pp. 1272--1286, 2020.

\bibitem{goury2018fast}
O.~Goury and C.~Duriez, ``Fast, generic, and reliable control and simulation of soft robots using model order reduction,'' \emph{IEEE Transactions on Robotics}, vol.~34, no.~6, pp. 1565--1576, 2018.

\bibitem{case2016sensor}
J.~C. Case, E.~L. White, and R.~K. Kramer, ``Sensor enabled closed-loop bending control of soft beams,'' \emph{Smart Materials and Structures}, vol.~25, no.~4, p. 045018, 2016.

\bibitem{zhao2021shape}
Q.~Zhao, J.~Lai, K.~Huang, X.~Hu, and H.~K. Chu, ``Shape estimation and control of a soft continuum robot under external payloads,'' \emph{IEEE/ASME Transactions on Mechatronics}, 2021.

\bibitem{hughes2021sensing}
J.~Hughes, F.~Stella, C.~D. Santina, and D.~Rus, ``Sensing soft robot shape using {IMU}s: An experimental investigation,'' in \emph{International Symposium on Experimental Robotics}.\hskip 1em plus 0.5em minus 0.4em\relax Springer, 2021, pp. 543--552.

\bibitem{wang2018toward}
H.~Wang, M.~Totaro, and L.~Beccai, ``Toward perceptive soft robots: Progress and challenges,'' \emph{Advanced Science}, vol.~5, no.~9, p. 1800541, 2018.

\bibitem{zhao2016optoelectronically}
H.~Zhao, K.~O’Brien, S.~Li, and R.~F. Shepherd, ``Optoelectronically innervated soft prosthetic hand via stretchable optical waveguides,'' \emph{Science robotics}, vol.~1, no.~1, p. eaai7529, 2016.

\bibitem{galloway2019fiber}
K.~C. Galloway, Y.~Chen, E.~Templeton, B.~Rife, I.~S. Godage, and E.~J. Barth, ``Fiber optic shape sensing for soft robotics,'' \emph{Soft robotics}, vol.~6, no.~5, pp. 671--684, 2019.

\bibitem{elgeneidy2018directly}
K.~Elgeneidy, G.~Neumann, M.~Jackson, and N.~Lohse, ``Directly printable flexible strain sensors for bending and contact feedback of soft actuators,'' \emph{Frontiers in Robotics and AI}, p.~2, 2018.

\bibitem{navarro2020model}
S.~E. Navarro, S.~Nagels, H.~Alagi, L.-M. Faller, O.~Goury, T.~Morales-Bieze, H.~Zangl, B.~Hein, R.~Ramakers, W.~Deferme, \emph{et~al.}, ``A model-based sensor fusion approach for force and shape estimation in soft robotics,'' \emph{IEEE Robotics and Automation Letters}, vol.~5, no.~4, pp. 5621--5628, 2020.

\bibitem{sahu2022spring}
S.~K. Sahu, I.~Tamadon, B.~Rosa, P.~Renaud, and A.~Menciassi, ``A spring-based inductive sensor for soft and flexible robots,'' \emph{IEEE Sensors Journal}, vol.~22, no.~20, pp. 19\,931--19\,940, 2022.

\bibitem{alcaide2017design}
J.~O. Alcaide, L.~Pearson, and M.~E. Rentschler, ``Design, modeling and control of a sma-actuated biomimetic robot with novel functional skin,'' in \emph{2017 IEEE International Conference on Robotics and Automation (ICRA)}.\hskip 1em plus 0.5em minus 0.4em\relax IEEE, 2017, pp. 4338--4345.

\bibitem{hofer2021vision}
M.~Hofer, C.~Sferrazza, and R.~D’Andrea, ``A vision-based sensing approach for a spherical soft robotic arm,'' \emph{Frontiers in Robotics and AI}, vol.~8, p. 630935, 2021.

\bibitem{3D-Deformation-sensing}
R.~B.~N. Scharff, G.~Fang, Y.~Tian, J.~Wu, J.~M.~P. Geraedts, and C.~C. Wang, ``Sensing and reconstruction of 3-d deformation on pneumatic soft robots,'' \emph{IEEE/ASME Transactions on Mechatronics}, vol.~26, no.~4, pp. 1877--1885, 2021.

\bibitem{scharff2022rapid}
R.~B. Scharff, D.-J. Boonstra, L.~Willemet, X.~Lin, and M.~Wiertlewski, ``Rapid manufacturing of color-based hemispherical soft tactile fingertips,'' in \emph{2022 IEEE 5th International Conference on Soft Robotics (RoboSoft)}.\hskip 1em plus 0.5em minus 0.4em\relax IEEE, 2022, pp. 896--902.

\bibitem{baaij2023learning}
T.~Baaij, M.~K. Holkenborg, M.~St{\"o}lzle, D.~van~der Tuin, J.~Naaktgeboren, R.~Babu{\v{s}}ka, and C.~Della~Santina, ``Learning 3d shape proprioception for continuum soft robots with multiple magnetic sensors,'' \emph{Soft Matter}, 2023.

\bibitem{della2020improved}
C.~Della~Santina, A.~Bicchi, and D.~Rus, ``On an improved state parametrization for soft robots with piecewise constant curvature and its use in model based control,'' \emph{IEEE Robotics and Automation Letters}, vol.~5, no.~2, pp. 1001--1008, 2020.

\bibitem{felt2019inductance}
W.~Felt, M.~J. Telleria, T.~F. Allen, G.~Hein, J.~B. Pompa, K.~Albert, and C.~D. Remy, ``An inductance-based sensing system for bellows-driven continuum joints in soft robots,'' \emph{Autonomous robots}, vol.~43, no.~2, pp. 435--448, 2019.

\bibitem{thuruthel2022multimodel}
T.~G. Thuruthel and F.~Iida, ``Multimodel sensor fusion for learning rich models for interacting soft robots,'' \emph{arXiv preprint arXiv:2205.04202}, 2022.

\bibitem{fang2019vision}
G.~Fang, X.~Wang, K.~Wang, K.-H. Lee, J.~D. Ho, H.-C. Fu, D.~K.~C. Fu, and K.-W. Kwok, ``Vision-based online learning kinematic control for soft robots using local gaussian process regression,'' \emph{IEEE Robotics and Automation Letters}, vol.~4, no.~2, pp. 1194--1201, 2019.

\bibitem{thuruthel2018model}
T.~G. Thuruthel, E.~Falotico, F.~Renda, and C.~Laschi, ``Model-based reinforcement learning for closed-loop dynamic control of soft robotic manipulators,'' \emph{IEEE Transactions on Robotics}, vol.~35, no.~1, pp. 124--134, 2018.

\bibitem{brena2020choosing}
R.~F. Brena, A.~A. Aguileta, L.~A. Trejo, E.~Molino-Minero-Re, and O.~Mayora, ``Choosing the best sensor fusion method: A machine-learning approach,'' \emph{Sensors}, vol.~20, no.~8, p. 2350, 2020.

\bibitem{qiu2022multi}
S.~Qiu, H.~Zhao, N.~Jiang, Z.~Wang, L.~Liu, Y.~An, H.~Zhao, X.~Miao, R.~Liu, and G.~Fortino, ``Multi-sensor information fusion based on machine learning for real applications in human activity recognition: State-of-the-art and research challenges,'' \emph{Information Fusion}, vol.~80, pp. 241--265, 2022.

\bibitem{fang2022efficient}
G.~Fang, Y.~Tian, Z.-X. Yang, J.~M. Geraedts, and C.~C. Wang, ``Efficient jacobian-based inverse kinematics with sim-to-real transfer of soft robots by learning,'' \emph{IEEE/ASME Transactions on Mechatronics}, 2022.

\bibitem{bern2020soft}
J.~M. Bern, Y.~Schnider, P.~Banzet, N.~Kumar, and S.~Coros, ``Soft robot control with a learned differentiable model,'' in \emph{2020 3rd IEEE International Conference on Soft Robotics (RoboSoft)}.\hskip 1em plus 0.5em minus 0.4em\relax IEEE, 2020, pp. 417--423.

\bibitem{wang2022control}
J.~Wang and A.~Chortos, ``Control strategies for soft robot systems,'' \emph{Advanced Intelligent Systems}, vol.~4, no.~5, p. 2100165, 2022.

\bibitem{george2018control}
T.~George~Thuruthel, Y.~Ansari, E.~Falotico, and C.~Laschi, ``Control strategies for soft robotic manipulators: A survey,'' \emph{Soft robotics}, vol.~5, no.~2, pp. 149--163, 2018.

\bibitem{han2009pid}
J.~Han, ``From pid to active disturbance rejection control,'' \emph{IEEE transactions on Industrial Electronics}, vol.~56, no.~3, pp. 900--906, 2009.

\bibitem{gerboni2017feedback}
G.~Gerboni, A.~Diodato, G.~Ciuti, M.~Cianchetti, and A.~Menciassi, ``Feedback control of soft robot actuators via commercial flex bend sensors,'' \emph{IEEE/ASME Transactions on Mechatronics}, vol.~22, no.~4, pp. 1881--1888, 2017.

\bibitem{bieze2018finite}
T.~M. Bieze, F.~Largilliere, A.~Kruszewski, Z.~Zhang, R.~Merzouki, and C.~Duriez, ``Finite element method-based kinematics and closed-loop control of soft, continuum manipulators,'' \emph{Soft robotics}, vol.~5, no.~3, pp. 348--364, 2018.

\bibitem{azizkhani2022dynamic}
M.~Azizkhani, I.~S. Godage, and Y.~Chen, ``Dynamic control of soft robotic arm: A simulation study,'' \emph{IEEE Robotics and Automation Letters}, vol.~7, no.~2, pp. 3584--3591, 2022.

\bibitem{chu2022feedback}
X.~Chu, R.~Ng, H.~Wang, and K.~W.~S. Au, ``Feedback control for collision-free nonholonomic vehicle navigation on se (2) with null space circumvention,'' \emph{IEEE/ASME Transactions on Mechatronics}, vol.~27, no.~6, pp. 5594--5604, 2022.

\bibitem{Model-based-sensing}
Z.~Q. Tang, H.~L. Heung, K.~Y. Tong, and Z.~Li, ``Model-based online learning and adaptive control for a “human-wearable soft robot” integrated system,'' \emph{The International Journal of Robotics Research}, vol.~40, no.~1, pp. 256--276, 2021.

\bibitem{zughaibi2021fast}
J.~Zughaibi, M.~Hofer, and R.~D’Andrea, ``A fast and reliable pick-and-place application with a spherical soft robotic arm,'' in \emph{2021 IEEE 4th International Conference on Soft Robotics (RoboSoft)}.\hskip 1em plus 0.5em minus 0.4em\relax IEEE, 2021, pp. 599--606.

\bibitem{li2022equivalent}
S.~Li, A.~Kruszewski, T.-M. Guerra, and A.-T. Nguyen, ``Equivalent-input-disturbance-based dynamic tracking control for soft robots via reduced-order finite-element models,'' \emph{IEEE/ASME Transactions on Mechatronics}, vol.~27, no.~5, pp. 4078--4089, 2022.

\end{thebibliography}

\vspace{-10 mm}
\begin{IEEEbiography}[{\includegraphics[width=1in,height=1.25in,clip,keepaspectratio]{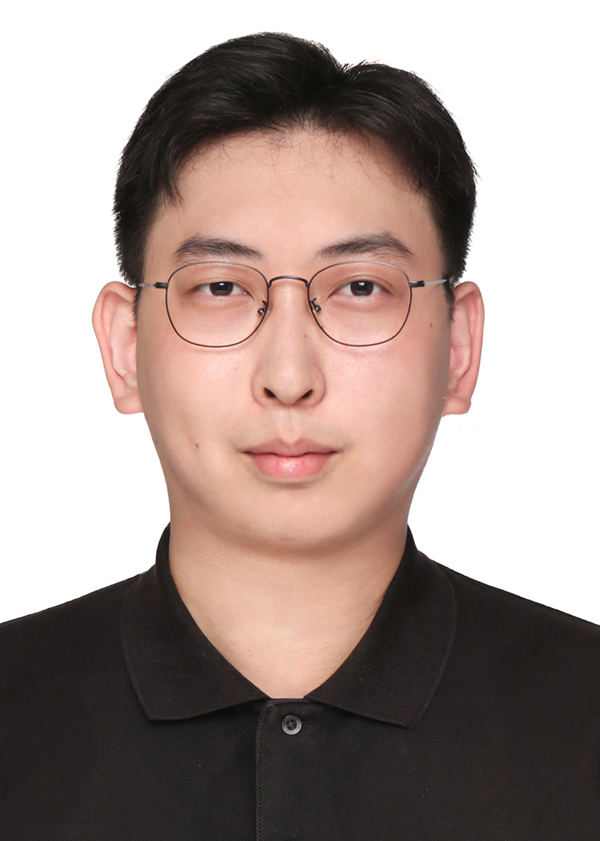}}]{Yinan Meng} received the Master's degree in Mechanical Engineering from the University of Southern California, Los Angeles, US, in 2018. He is working toward a Ph.D. with the Digital Manufacturing Lab, School of Engineering, The University of Manchester, Manchester, U.K. His research interest includes proprioception and control of soft robots.
\end{IEEEbiography}
\vspace{-10 mm}
\begin{IEEEbiography}[{\includegraphics[width=1in,height=1.25in,clip,keepaspectratio]{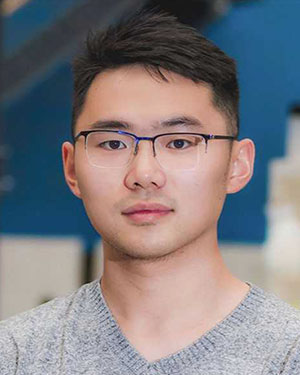}}]{Guoxin Fang}
(Member, IEEE) received the B.E. degree in mechanical engineering from the Beijing Institute of Technology, Beijing, China, in 2016 and the Ph.D. degree in advanced manufacturing from the Delft University of Technology, Delft, The Netherlands in 2022. He is now Research Associate with the Department of Mechanical, Aerospace and Civil Engineering, The University of Manchester, Manchester, U.K. His research interests include computational design, digital fabrication, and robotics.
\end{IEEEbiography}
\vspace{-10 mm}
\begin{IEEEbiography}[{\includegraphics[width=1in,height=1.25in,clip,keepaspectratio]{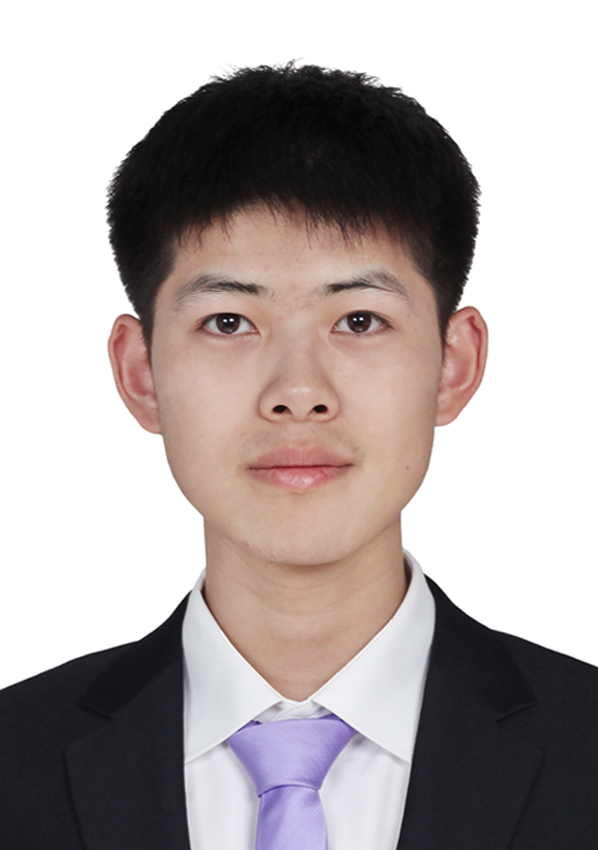}}]{Jiong Yang} received the Master's degree of Mechanical Engineering in the University of Manchester, Manchester, UK, in 2018.  He is currently working on a PhD project in Advanced Manufacturing Group, Department of Mechanical Aerospace and Civil Engineering, School of Engineering, the University of Manchester. He researches on novel additive manufacturing (AM) methods in tissue engineering and bioengineering.
\end{IEEEbiography}
\vspace{-10 mm}
\begin{IEEEbiography}[{\includegraphics[width=1in,height=1.25in,clip,keepaspectratio]{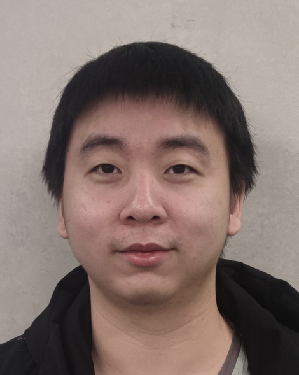}}]{Yuhu Guo} received the bachelor's degree in software engineering from Xiamen University in 2021. He is currently studying for an MPhil degree in the Smart Manufacturing Group of the University of Manchester, UK. His current work is at the intersection of digital fabrication, computational design, soft robotics, and machine learning.
\end{IEEEbiography}
\vspace{-10 mm}
\begin{IEEEbiography}[{\includegraphics[width=1in,height=1.25in,clip,keepaspectratio]{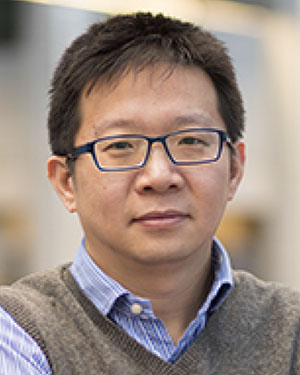}}]{Charlie C.L. Wang} (Senior Member, IEEE) received the B.Eng. degree in mechatronics engineering from the Huazhong University of Science and Technology, China in 1998, and Ph.D. degrees in mechanical engineering from The Hong Kong University of Science and Technology in 2002. He is currently Professor and Chair of Smart Manufacturing with the University of Manchester, U.K. His research interests include digital manufacturing, computational design, soft Robotics, mass personalization, and geometric computing. He was elected as a Fellow of the American Society of Mechanical Engineers, in 2013.
\end{IEEEbiography}

\vfill


\end{document}